\documentclass[journal]{IEEEtran}
\usepackage{lipsum} 
\usepackage[tight,footnotesize]{subfigure}
\usepackage{footnote}
\usepackage{enumitem} 
\usepackage{float}
\usepackage{multicol,multienum}
\usepackage{breqn}
\usepackage{stfloats}
\usepackage{multirow}
\usepackage{diagbox}
\usepackage{hyperref}
\usepackage{bm}
\usepackage{cite}
\usepackage{graphicx}
\usepackage{tabularx}
\usepackage[ruled,vlined]{algorithm2e} 
\usepackage{algorithmic}
\usepackage{booktabs}

\newtheorem{definition}{Definition}

\usepackage[tight,footnotesize]{subfigure}
\usepackage{url}
\usepackage{doi}
\usepackage{amssymb}
\usepackage{amsmath}
\usepackage{color}
\usepackage{footnote}
\usepackage{changepage}
\usepackage{threeparttable}
\usepackage[defaultcolor=red,addedmarkup=uline]{changes}
\usepackage{soul}
\usepackage{makecell}
\usepackage{siunitx}

\usepackage[defaultcolor=red,addedmarkup=uline]{changes}
\setcommentmarkup{\footnote{\color{authorcolor} #1}}

\usepackage{soul}
\soulregister \cite7
\soulregister\ref7
\soulregister\pageref7

\ifCLASSINFOpdf

\else

\fi

\begin{document}\sloppy

\title{Traffic Prediction with Transfer Learning: A Mutual Information-based Approach}

\author{Yunjie~Huang, 

}


\maketitle

\begin{abstract}
In modern traffic management, one of the most essential yet challenging tasks is accurately and timely predicting traffic. It has been well investigated and examined that deep learning-based Spatio-temporal models have an edge when exploiting Spatio-temporal relationships in traffic data. Typically, data-driven models require vast volumes of data, but gathering data in small cities can be difficult owing to constraints such as equipment deployment and maintenance costs. To resolve this problem, we propose TrafficTL, a cross-city traffic prediction approach that uses big data from other cities to aid data-scarce cities in traffic prediction. Utilizing a periodicity-based transfer paradigm, it identifies data similarity and reduces negative transfer caused by the disparity between two data distributions from distant cities. In addition, the suggested method employs graph reconstruction techniques to rectify defects in data from small data cities. TrafficTL is evaluated by comprehensive case studies on three real-world datasets and outperforms the state-of-the-art baseline by around 8 to 25 percent.
\end{abstract}


\IEEEpeerreviewmaketitle



\section{Introduction}\label{sec:intro}
\IEEEPARstart{R}{eliable} traffic forecasting is considered an indispensable component in modern smart city construction, which can benefit tasks such as route planning, traffic control, and traffic network management \cite{Guo2021LearningDA,Li2020ShorttermTP, Yin2020ACS} in the Intelligent Transportation Systems (ITSs). A number of studies have investigated traffic prediction for decades. Autoregressive integrated moving average (ARIMA) \cite{ARIMA:Ahmed1979ANALYSISOF} and Kalman filter have been widely used for time-series data in traffic prediction problems. However, these methods use only historical observations from a single site to make predictions of future speed at that location. As technology advances, data collection devices' geographical information, i.e., longitude and latitude, are easily accessible for constructing spatial relationships among the raw data. Convolutional Neural Networks (CNNs) are applied to capture spatial features, considering that traffic states in nearby areas are influencing each other. In addition, taking the historical observation value of time-series into account,  Recurrent Neural Networks (RNN) are popular in capturing temporal features within traffic data.

In recent years, traffic prediction methods, especially graph-based methods, have attracted remarkable attention from the intelligent transportation community \cite{Zhao2020TGCNAT}. Benefiting from the fact that road networks inherently resemble graph structures, i.e., data collection devices on roads are represented as nodes, and the connection between the roads are represented as edges, Graph Neural Networks (GNNs) are intuitively more suitable than CNNs to capture the non-Euclidean spatial features from the road network \cite{yu2018spatio}.
%
Representative methods including Temporal Graph Convolutional Network (T-GCN) model \cite{Zhao2020TGCNAT}, Spatio-Temporal Graph Convolutional Networks (STGCN) \cite{yu2018spatio}, Diffusion Convolutional Recurrent Neural Network (DCRNN) \cite{Li2018DiffusionCR}  
exploit GNNs to model the spatial information of road network and furthermore embed it by the fine-grained high-dimensional features, and achieving satisfactory results.

Although graph-based methods have brought strides in traffic prediction, these methods ignore a practical problem of this ITS application: the difficulty in data acquisition.
Despite the emergence of the industrial internet of things and embedded computing devices, massive amounts of traffic data are collected by interconnected stationery or dynamic traffic devices (e.g., loop detectors \cite{loop1:Huang2014DeepAF,loop2:Lv2015TrafficFP}, radio frequency identification detectors \cite{RFID:Chao2014AnIT}, and traffic cameras \cite{video:Xia2016TowardsIQ}).
%
However, given the high maintenance costs, medium and small cities are unable to deploy or maintain a large number of traffic sensors in the long term to collect sufficient available traffic data. Consequently, these cities may lack the methods of accessing data or the correct prior knowledge to guide the construction of smart cities \cite{Wang2021EstimatingTF}, in particular ITS. Simultaneously, a lack of smart city construction precludes the acquisition of more data.
This phenomenon is also known as the ``cold start'' problem.
It also leads to a data imbalance in traffic data collection in different cities. 
For instance, smaller cities are more challenging to acquire data than larger cities, and they have relatively rare knowledge of their historical patterns. On the other hand, megacities with improving infrastructure can conveniently collect traffic data from more connected vehicles.
Consequently, devising a reliable traffic forecasting algorithm upon scarce and imbalanced data is essential for small and medium cities.
The recent transfer learning technique can be a feasible solution to achieve this objective.
The core idea of transfer learning for traffic prediction is to use large amounts of data from other cities to provide intelligent predictions for cities with scarce historical data \cite{Transfer:yangPan2010ASO}. It has preliminary applications in other data analytic services, e.g., air quality detection \cite{Wei2016TransferKB}, ride-sharing detection \cite{Wang2019RidesharingCD}, and cross-city chain-store site-selection recommendation \cite{Guo2017CityTransferTI}. 

In the context of transfer learning, we regard the city that provides a significant volume of data as the \textit{source} city.
In the meantime, the \textit{target} city refers to a city where data is scarce.
%
In transferable traffic forecasting tasks, \cite{Li2020ShorttermTP} trains a model directly with the large amount of source data collected by the source city and then fine-tunes it with a small amount of target data from the target city in the model's last few layers.
\cite{Xu2016CrossregionTP} used support vector machine, and \cite{Lin2018TransferLF} applied dynamic time wrapping to assist cross-city speed prediction. Others introduced graph clustering to learn data transfer methods \cite{Mallick2021TransferLW}. 
However, the research above faces notable challenges: 1) Uneven data distribution in different cities may easily cause negative transfer of models in transfer learning, which refers to the phenomenon that knowledge learned on the source domain\footnote{``Domain'' in transfer learning consists of two parts: the feature space of the city's data and its marginal probability distribution. Typically, one city is considered to be one domain.} has a negative effect on target domain feature extraction \cite{Pan2010ASO}; 
2) Although the core discipline of transfer learning is to share the same knowledge in two similar domains, sharing incorrect prior knowledge may lead to significant error accumulation \cite{Bai2020AGCRN}.
To summarize, existing transfer learning-based traffic speed forecasting approaches mainly share the following drawbacks:

\begin{itemize}
    \item \textbf{Data distribution differences:}
    While transfer learning is to learn from the source domain and transfer knowledge to the target domain, 
    in almost all cases, the source and target domains follow two different distributions. The difference in data distribution has a non-negligible adverse impact on the effectiveness of transfer (negative transfer).
    In addition, \cite{Du2021AdaRNNAL} points out that even time-series from the same data source have uneven data distribution in different time slots.
    Therefore, avoiding negative transfer is a significant challenge in this problem. 
    
    \item \textbf{prior knowledge error:}
	Existing traffic data often record the GPS geographic location of detectors, allowing GNN-based methods to capture features on the non-Euclidean spatial structure constructed by location information of traffic data.
    This approach requires predefined interconnection information developed by similarity or metric distance, which requires significant domain knowledge \cite{Geng2019SpatiotemporalMC}. 
    Furthermore, graphs generated in this way are usually counter-intuitive and incomplete, contributing irrelevantly to the target task \cite{Bai2020AGCRN}.
    In addition, the data collection in the target cities is limited for the high costs of deploying and maintaining facilities. Consequently, there are concerns such as missing data from traffic sensors and incorrectly designed adjacency matrix, indicating that prior knowledge is not always reliable.
    
    \item \textbf{Indistinguishable parameter sharing}:
    Although deep learning-based traffic prediction algorithms have shown promising results, they tend to accentuate prominent properties in the dataset. 
    The data from the dataset follow the same distribution, making traffic prediction possible within one dataset.
    However, the shared parameters make current methods perform poorly in capturing fine-grained data features if transferred to another city \cite{Bai2020AGCRN}.
    As mentioned above, considering the different distributions of two domains, directly transferring the model from the source city to the target may inevitably result in inaccurate predictions. 
    Furthermore, the error caused by such knowledge gaps accumulates during training.
\end{itemize}

To address these challenges, we design a graph-based traffic prediction framework with transfer learning.
In particular, the main highlights of the proposed approach are summarized as follows:

\begin{figure}[!t]
	\centering
	\includegraphics[scale=0.33]{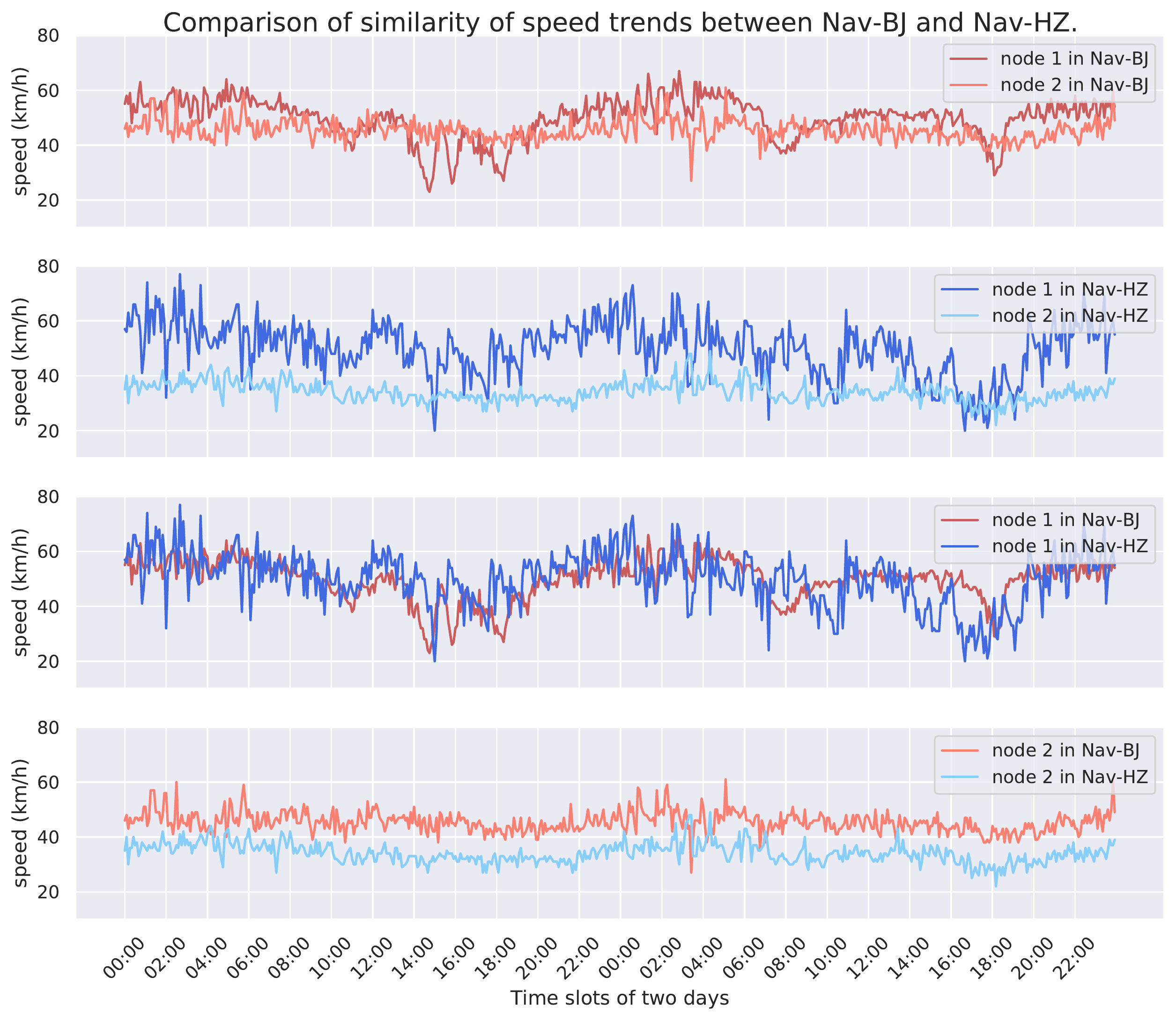}
	\caption{Red lines represents two Nav-BJ roads and blue lines are two Nav-HZ roads, which come from different datasets. The figure shows a case where roads in the same city can have different daily traffic trend, and ones from different cities can be similar.
}
	\label{fig:temporalsim}
  \end{figure}

\begin{itemize}
	\item \textbf{Temporal Cluster Block:} 
	Temporal periodicity has been proposed early on traffic tasks but is rarely applied to transfer learning \cite{Zou2015HybridSF}. 
	The	graph structure's properties may differ for two data with divergent distributions. However, traffic speed trends over time can be similar in different cities, as shown in Fig. \ref{fig:temporalsim}. Such common features give us the idea of using transfer learning methods for traffic prediction for small city governors. However, clustering of two differently distributed data in transfer learning with conventional methods may produce degenerate solutions. Therefore, we propose a temporal clustering block (TCB) that uses mutual information to better cluster road segments with similar temporal properties.
	
	\item \textbf{Graph Reconstruction Block:}  
	For the second challenge, we propose a graph reconstruction block (GRB) that generates a new graph of the clustered data produced by TCB. The clustered data includes a subset of the source and target data.
	This block uses the clustered data without raw graph structure information to generate a new adjacency matrix. 
	The newly generated graph better focuses on trend-categorically consistent data, i.e., data with stronger speed-time trend relevance, which is helpful to retain similar trend information during the transfer.

	\item \textbf{Ensemble model:} 
	To address the third challenge (indistinguishable parameter sharing), we propose an ensemble model particularly focusing on local parameter sharing.
	We feed multiple batches of clustered data into different sub-models for training, reducing the data divergence among the sub-models. 
	Consequently, the more data similarity, the better the transfer efficiency, notably reducing negative transfer influence.
\end{itemize}

The rest of this paper is organized as follows. 
We first review the background of traffic prediction, time-series cluster, and transfer learning in Sec. \ref{section:II}. Then, we present the preliminaries in Sec. \ref{section:III}. Sec. \ref{section:IV} elaborates on the proposed system. We conduct case studies and provide analytical discussions in Sec. \ref{section:V}. Finally, we conclude this paper in Sec. \ref{section:VI}.

\section{Related Work}\label{section:II}
In this section, we review previous efforts on traffic prediction. Transfer learning and time-series unsupervised classification-related works are also presented.
\subsection{Traffic Prediction}

Traffic prediction has been extensively studied in past decades. Some statistical methods, such as HA \cite{HA1:Jeffery1987ELECTRONICRG} and ARIMA \cite{ARIMA:Ahmed1979ANALYSISOF}, are broadly adopted in the time-series community. But these methods only consider temporal information, neglecting the impact of adjoining roadways on each other.
Deep learning approaches show superior performance in capturing spatial-temporal correlations and non-linear relations, compared to the statistical methods above and machine learning techniques such as support vector regression (SVR) \cite{SVR:Ding2002TrafficFT} and K-nearest neighbor (KNN).
Recurrent Neural Network (RNN) is well-recognized for modeling time-series by extracting information from past instances to capture temporal features.
Yet it suffers from the gradient vanishing and exploding problems \cite{RNNvanish}, \cite{RNNvanish2}.
%
The development of Long Short-Term Memory (LSTM) and Gated Recurrent Unit (GRU) \cite{GRU:Wu2018AHD, LSTMfottraffic, CNNLSTMVMittal} has advanced the capabilities of Recurrent Neural Networks (RNNs) for processing time-series data. These techniques have evolved RNNs into a new generation of models for handling time-dependent data.
They address the shortcomings of RNN and GRU, which further significantly alleviates the computation burden by reducing the total number of parameters learned from raw data.

In addition to RNNs and their variations, CNNs are another type of deep learning approach that is frequently utilized to incorporate spatial contextual information for traffic prediction tasks \cite{CNN:Lv2015TrafficFP, CNNLSTM:Zhang2017DeepSR, CNNLSTMVMittal}.
In vanilla CNN-based traffic predictors, the geographic map is intuitively divided into square cells to form a grid. Each cell's incoming and outgoing traffic flow data are aggregated as the ground truth for prediction \cite{CNN:Ma2015LargeScaleTN}. 
However, this approach destroys the non-Euclidean structure of the road network. 
As a recent alternative, graph-based neural networks provide a better choice to preserve the connectivity and adjacency information of traffic networks \cite{Zhao2020TGCNAT,yu2021long,fastgnn,james2022graph,ASTGCN}.
Specifically, Graph Convolutional Networks (GCNs) utilize principles of spectral graph theory to extend the concept of convolutional operations to non-Euclidean domains, as demonstrated in \cite{GCNshuguangyang, GCNxixiong}. These networks have been shown to be effective in a variety of tasks, including those involving graph-structured data.
For example, \cite{Zhao2020TGCNAT} presented T-GCN and applied GCN to learn the topology of road networks and GRU to learn the dynamical changes of traffic flow. 
\cite{ASTGCN} proposed an Attention-based Spatial-Temporal Graph Convolutional Network (ASTGCN), which incorporates an attention mechanism to better model spatial-temporal correlation. Furthermore, \cite{Li2018DiffusionCR} extends the graph convolution operator by introducing diffusion convolution and proposes a deep learning framework named DCRNN for traffic flow prediction. It uses bi-directional random-walk on graphs to capture spatial correlation and encoder-decoder architecture with GRU to capture temporal correlation.
\cite{Dynamic-graph} uses a dynamic graph to improve traffic flow prediction. \cite{Traffic-flow} proposes a new graph neural network layer with an attention mechanism to better aggregate traffic flows from neighboring roads. 

The aforementioned methods provide extra spatial information to traffic prediction while improving the prediction's overall performance.
However, the road structures captured by these methods are all predicated based on acquiring complete, high-quality, and large volumes of traffic data. 
For cities that lack accurate or sufficient historical data, typical traffic predictors are insufficient to generate high-quality forecasts. Alternative solutions are required in this context. 


\subsection{Transfer Learning for Traffic Forecasting Problems}
Research on building traffic forecasting models has significantly progressed and been applied in practical applications. 
However, these outstanding results are mainly produced on large-scale datasets, making it difficult to train a model with limited data. 
Concerning this fact, methods like RegionTrans \cite{Wang2019CrossCityTL}, and MetaST \cite{Yao2019LearningFM} try to tackle the data scarcity problems in traffic forecasting by applying transfer learning techniques. 
These two works focus on transferring knowledge by building a region-matching and co-training principle between cities. 
RegionTrans \cite{Wang2019CrossCityTL} transfers knowledge between cities via regional similarity regularization. 
MetaST \cite{Yao2019LearningFM} uses a vector array named ST-mem to store the meta-information of source cites' region clusters. 
It subsequently transfers knowledge between domains by matching the target city's region to ST-mem with an attention mechanism-based approach.
Nonetheless, either approach relies heavily on the quality of initial road clustering: anomalous ones may lead to negative transferability. 
%
Both methods are built based on parameter sharing, which shares network parameters during model training and transferring.
They optimize the model by calculating the losses between the prediction and ground truth on different matching regions and combining them with pre-defined coefficients. 
However, they ignore the inductive sharing bias between the source and target domain, i.e., the gradients computed on different sub-datasets clustered by region matching are likely to cause conflicts in the gradient descent direction, thereby undermining the overall performance of the model. 


Considering the defect of the aforementioned ``global'' parameter sharing scheme, local parameter sharing is a worthy strategy to reduce the negative transfer between models. TL\_DCRNN \cite{Mallick2021TransferLW} reduce information sharing on graphs with fewer correlations by graph partitioning. However, this graph partitioning approach requires that the sub-graph sizes be aligned. Otherwise, the miss-aligned parts will be filled with zero values, adversely affecting learning performance. 
Moreover, as the graph partitioning operation is required on both the source and target datasets, the accuracy of traffic graphs becomes a crucial basis that influences the transfer efficiency.

In this work, we follow this idea of sharing parameters in local regions. This requires clustering methods to relax the strong dependence on graph structure. Following, we briefly introduce the major existing time-series clustering methods.

\subsection{Time-series Clustering}
Time-series clustering methods are typically classified into three categories: \textit{Whole time-series clustering}, \textit{Subsequence time-series clustering}, and \textit{Time points clustering}. We focus on the \textit{Whole time-series clustering} approach for its clear correlation to this work. 
Interested readers are referred to \cite{Aghabozorgi2015TimeseriesC} for a more comprehensive overview of the clustering method.

\textit{Whole time-series clustering} refers to clustering approaches on discrete objects, where the objects in our work are the discrete nodes with temporal properties.
Generally, these clustering methods can be further divided into the following groups: partitioning, hierarchical, grid-based, density-based, model-based, and multi-step clustering \cite{Aghabozorgi2015TimeseriesC}.
%
%

The most common partitioning method divides the $n$ unlabeled nodes into $k$ groups, each containing at least one node from the dataset. K-means \cite{macqueen1967some} and K-Medoids \cite{Hadi1991FindingGI} are two widely-recognized representatives.
However, they face the challenge of determining the number of clusters $k$ in advance, which is difficult in practice \cite{Hadi1991FindingGI}. While the hierarchical analysis does not have this trouble,
it is restricted to small datasets due to the quadratic computing complexity \cite{hierarchicaldrawback}. 
At the same time, the grid-based method is either too slow or too imprecise \cite{Ester1996ADA}. 
And the density-based way is not widely employed in time-series data clustering due to its high complexity \cite{Aghabozorgi2015TimeseriesC}.
Model-based methods attempt to recover the original model from a set of data.
Representative models are Markov chain \cite{Markov}, Neural Network \cite{SOM}, and Gaussian mixture class models \cite{gaussian}, etc.
Previously, neural networks were considered a less favored technique due to their prolonged execution time and small data volume capacity.
Nevertheless, the recent development of computing facilities and deep learning techniques assist neural networks in analyzing larger amounts of data with improved time and accuracy performance. 
This method is becoming more prevalent than ever in time-series clustering.
Finally, the multi-step method integrates multiple methods introduced above, aiming at their respective advantages. In addition, \cite{High-per} proposes calculating the temporal similarity within a fixed time interval and treating each road section as a single cluster merging the most similar sections bottom-up to achieve clustering. \cite{Research-on} uses a hierarchical clustering algorithm with Pearson correlation coefficients, again by continuously merging similar road sections for clustering. \cite{Urban-traffic} uses Gaussian mixture model clustering. The first two, taken together, use a hierarchical clustering approach to merge bottom-up sections with similar distance similarity functions. Still, this approach has vague termination conditions, making the cluster formation difficult to change. In contrast, the third uses a GMM approach to clustering using probability, which may result in a single data point belonging to multiple clusters, not the effect we want to achieve in our task. 
\section{Transferable Traffic Prediction}\label{section:III}

This section introduces the preliminaries involved in transfer prediction, including the definition of the traffic network, transfer learning in traffic data, and the definition of traffic prediction with transfer learning.

\subsection{Traffic Network}
We define each traffic road network as a graph $G = (\mathcal{V}, \mathcal{E}, \bm{A})$, where $\mathcal{V}$ is a set of $N$ nodes  (i.e., traffic sensors in the city's roads.); $\mathcal{E}$ is a set of edges, presenting the connectivity among $\mathcal{V}$; $\bm{A} \in \mathbb{B}^{N \times N}$ is the adjacent matrix of $G$.
%
%
\begin{figure}[!t]
	\centering
	\includegraphics[scale=0.5]{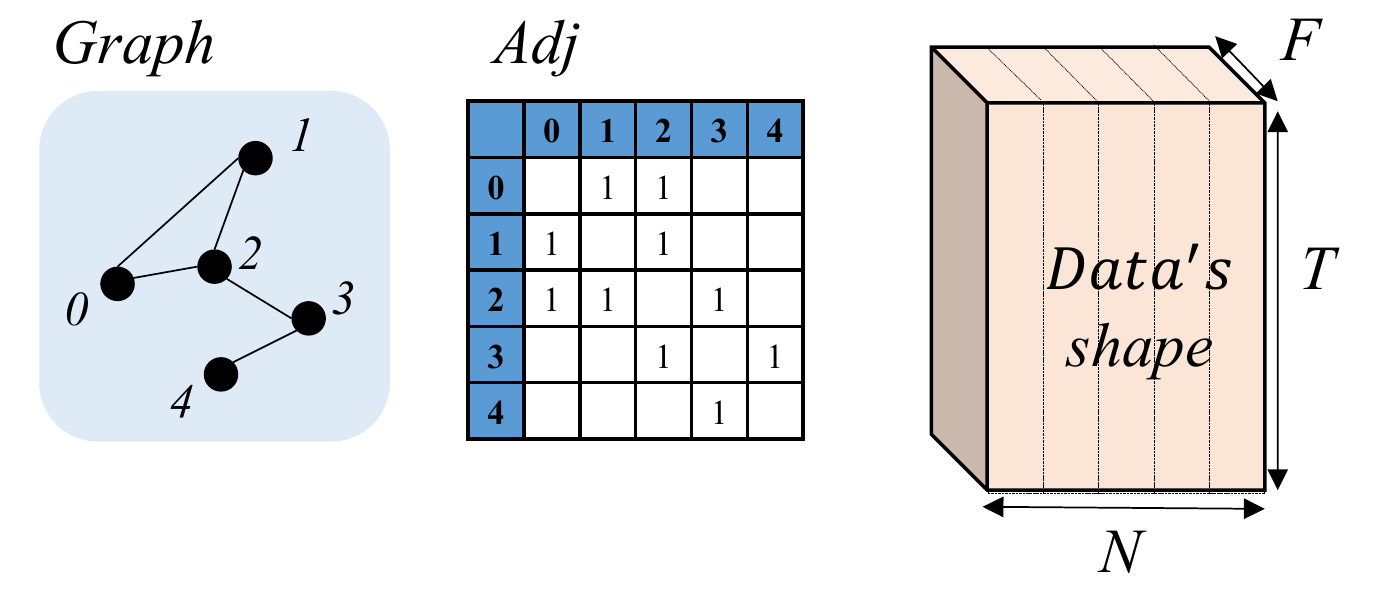}
	\caption{Graph construction process and the resulting feature tensor.}
	\label{fig:datadescrib}
  \end{figure}
%
Fig. \ref{fig:datadescrib} gives a typical example of this graph construction scheme and the resulting tensor formulation.
If node 1 has a connection to node 2, $A_{12} = A_{21} = 1$. 
Note that we reconstruct the adjacency matrix in Sec. \ref{sec:GRB} for the proposed approach to capture the data correlation when historical data is insufficient to build the nodal connectivity.

\begin{figure}[!t]
	\centering
	\includegraphics[scale=0.3]{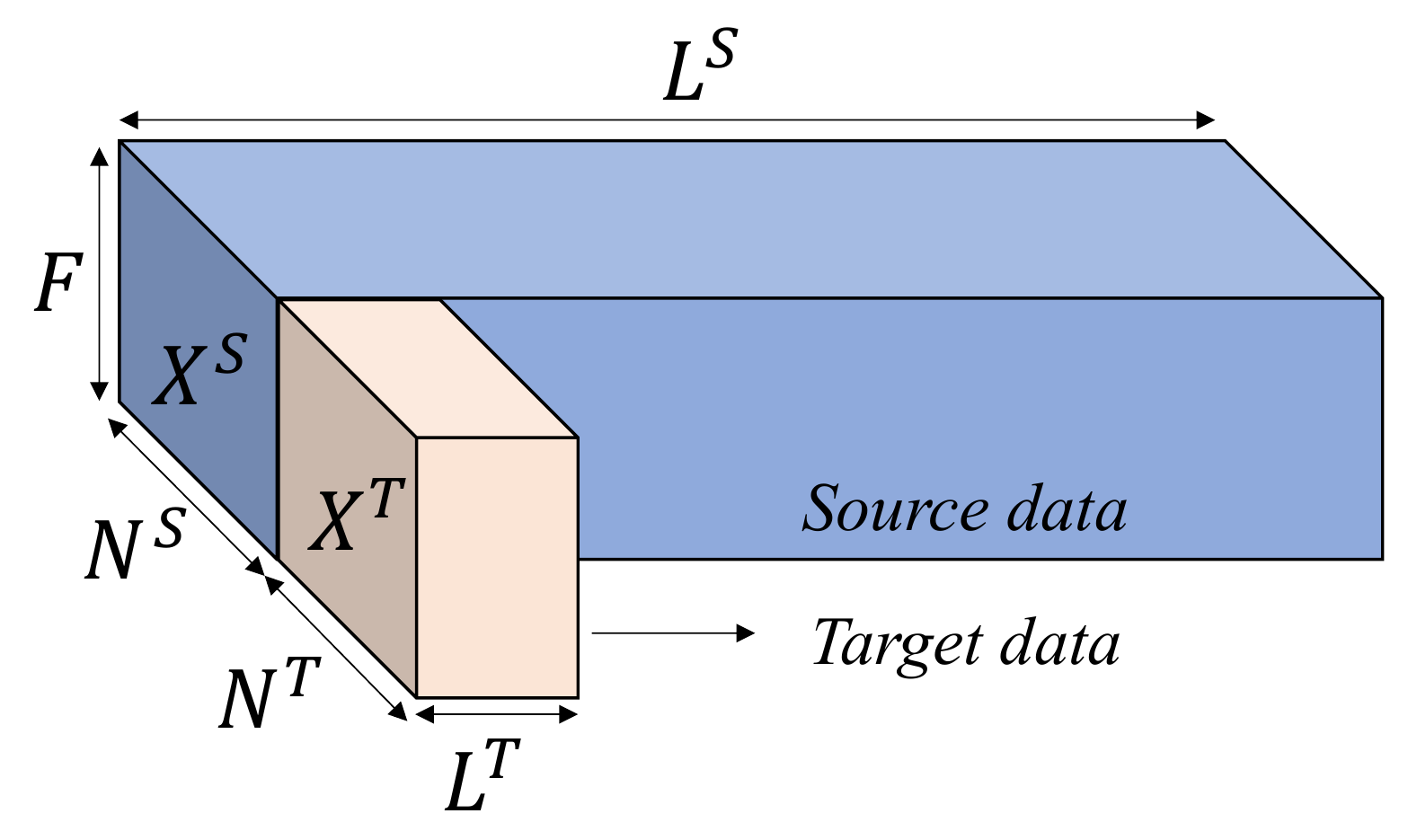}
	\caption{Data of the source and target cities in transfer learning.}
	\label{fig:transferdata}
  \end{figure}

\begin{figure*}[!ht]
 \centering
 \includegraphics[scale=0.68]{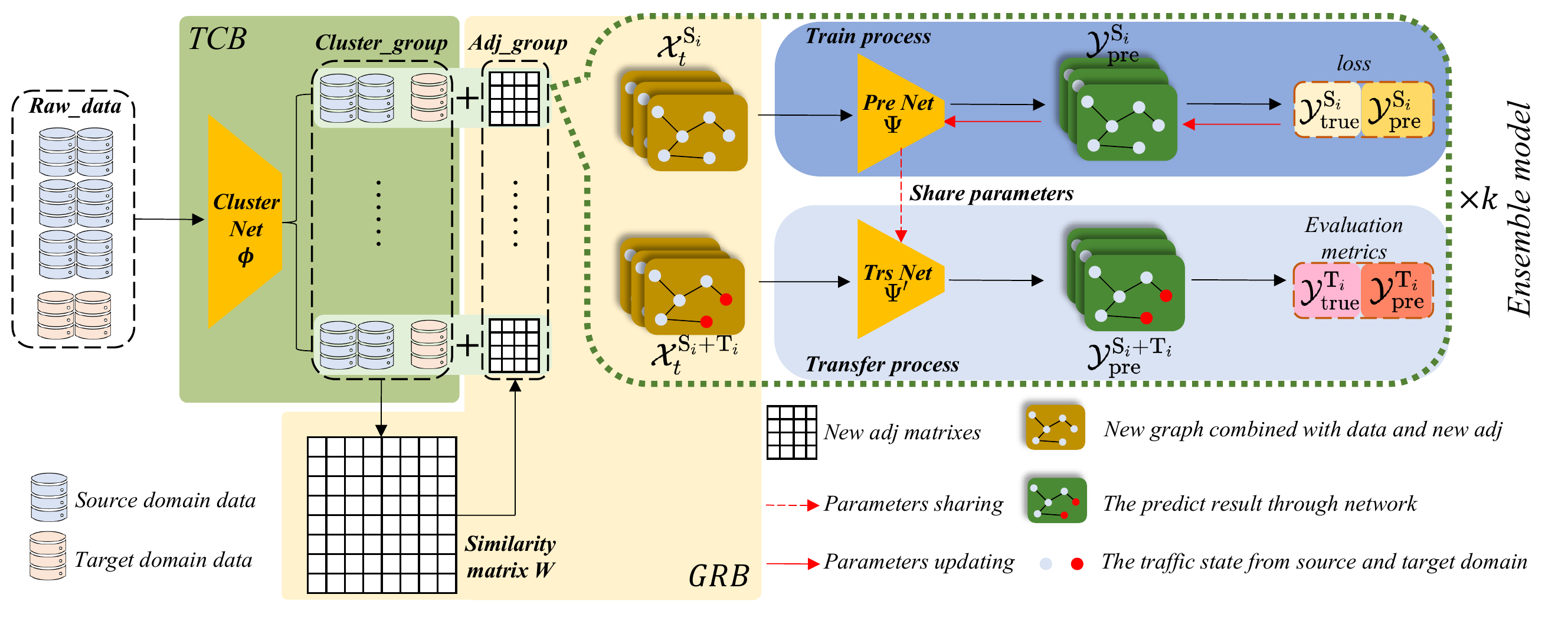}
 \caption{Overview of the proposed traffic prediction framework with transfer learning. The TCB module clusters data with similar periodicity in the source and target city. GRB reconstructs the traffic graph with the similarity matrix $W$. The training process uses the source data for Spatio-temporal information extraction, and the transfer process replaces part of the source data with target data to get the prediction results.}
  \label{fig:overview}
\end{figure*}
  
\subsection{Transfer Learning in Traffic Network}\label{sec:TL}
Transfer learning is the knowledge transfer between the source and the target domain. 
Let superscripts $\mathcal{S}$ and $\mathcal{T}$ represent the data of the source and target cities, respectively. Subsequently, we denote the road network graph of the source and target cities as $G^{\mathcal{S}} = (\mathcal{V}^{\mathcal{S}}, \mathcal{E}^{\mathcal{S}}, \bm{A}^{\mathcal{S}})$, $G^{\mathcal{T}} = (\mathcal{V}^{\mathcal{T}}, \mathcal{E}^{\mathcal{T}}, \bm{A}^{\mathcal{T}})$ respectively.
Note that the number of nodes in the two cities can be different. Fig. \ref{fig:transferdata} gives an illustration to the shape of source and target city data, namely, raw source data $X^{\mathcal{S}} \in \mathbb{R}^{L^{\mathcal{S}} \times N^{\mathcal{S}} \times F}$ and raw target data $X^{\mathcal{T}} \in \mathbb{R}^{L^{\mathcal{T}} \times N^{\mathcal{T}} \times F}$, where $N^{\mathcal{S}} \neq N^{\mathcal{T}}$ and the number of nodes in both cities, $L^{\mathcal{S}} \gg L^{\mathcal{T}}$ are the length of time in the raw data from both cities, and $F$ is the number of traffic features. In our work, we primarily investigate the traffic speed data transferability, i.e., $F=1$.





\subsection{Problem Definition} \label{sec:pro-def}
The primary objective of transferable traffic prediction is to predict future traffic data in data-scarce cities with the transferred traffic knowledge from data-rich cities. In this work, we integrate the data from the target city $X^{\mathcal{T}}$ and the source city $X^{\mathcal{S}}$ in order to obtain \textit{trend-similar} node clusters $\mathcal{C} = (C_{1}, C_{2}, \cdots, C_{c})$ to facilitate the transfer, where $c$ is the number of clusters. Each cluster $C_{i}$ includes a subset $X^{\mathcal{S}} \in \mathbb{R}^{L^{\mathcal{S}} \times N^{\mathcal{S}}_{i} \times F}$, and $X^{\mathcal{T}} \in \mathbb{R}^{L^{\mathcal{T}} \times N^{\mathcal{T}}_{i} \times F}$ in the source and the target cities, respectively, where $\sum_{i=1}^{c} N^{\mathcal{S}}_{i} = N^{\mathcal{S}}$ and $\sum_{i=1}^{c} N^{\mathcal{T}}_{i} = N^{\mathcal{T}}$ are the respective nodes in all clusters.

For each cluster $C_{i}$, the $H$ historical traffic observations on graph $G^{\mathcal{S}_i}$ of the source city at time $t$ is denoted as $\mathcal{X}_{t}^{\mathcal{S}_{i}} = (x^{\mathcal{S}_{i}}_{t-H},\cdots, x^{\mathcal{S}_{i}}_{t-2},  x^{\mathcal{S}_{i}}_{t-1}) \in \mathbb{R}^{H \times N^{\mathcal{S}}_{i} \times F}$.
Each $x^{\mathcal{S}_i}_t \in \mathbb{R}^{1 \times N^{\mathcal{S}}_i \times F}$ represents a traffic state on sub-graph $G^{\mathcal{S}_i}$ at time $t$. The aggregated $\mathcal{X}^{\mathcal{S}_i}_t \in \mathbb{R}^{H \times N^{\mathcal{S}}_i \times F}$ is a set of traffic states from time $t-H$ to time $t-1$.
We apply $\mathcal{X}_{t}^{\mathcal{S}_{i}}$ to train a model for predicting the traffic speed of the next $Q$ time instances $\mathcal{Y}_{t}^{\mathcal{S}_{i}} = (x^{\mathcal{S}_{i}}_{t}, x^{\mathcal{S}_{i}}_{t+1}, \cdots, x^{\mathcal{S}_{i}}_{t+Q-1}) \in \mathbb{R}^{(Q) \times N^{\mathcal{S}}_{i} \times F}$. 
In the transferring process, we combine $\mathcal{X}_{t}^{\mathcal{S}_{i}}$ and $\mathcal{X}_{t}^{\mathcal{T}_{i}}$ to predict the next $Q$ values of the target city $\mathcal{Y}^{\mathcal{T}_{i}}_{t} = (x^{\mathcal{T}_{i}}_{t}, x^{\mathcal{T}_{i}}_{t+1}, \cdots, x^{\mathcal{T}_{i}}_{t+Q-1}) \in \mathbb{R}^{Q \times N^{\mathcal{T}}_{i} \times F}$.
Finally, the following $Q$-step predictions on all nodes of the target city at time $t$ are derived by the results from all clusters, i.e., 
$\mathcal{Y}^{\mathcal{T}}_{t} = \{ \mathcal{Y}^{\mathcal{T}_i}_t \in \mathbb{R}^{Q \times N^{\mathcal{T}}_i \times F}\}_i$.

\section{TrafficTL}\label{section:IV}

This section proposes a novel transferable traffic prediction model named TrafficTL to address the aforementioned challenges in Sec. \ref{sec:intro}. 
As depicted in Fig. \ref{fig:overview}, we first input the integrated source and target city data into the proposed clustering block (TCB), which is used to aggregate nodes with similar temporal trends.
With TCB, a group of clusters can be obtained, each containing a portion of nodes with a similar speed-trend in the source and target city.
Subsequently, we use a novel graph relation mining (GRB) module for each cluster to calculate the similarities among nodes in the cluster and obtain a similarity matrix $\bm{W}$, which is used as the adjacency matrix $\bm{A}$.
Finally, the adjacency matrix calculated by GRB and the source city data are adopted in an ensemble module for each cluster to train the network $\Psi$. 
During the transfer, We replace a percentage $\beta$ of the input data in the training network with nodes from the target city.
The training and transferable network share parameters to obtain the prediction results from the mixed cities' data.
We elaborate on the three key building blocks in the following sub-sections.

\subsection{Temporal Cluster Block}\label{sec:TCB}

\begin{figure}[!t]
	\centering
	\includegraphics[scale=0.4]{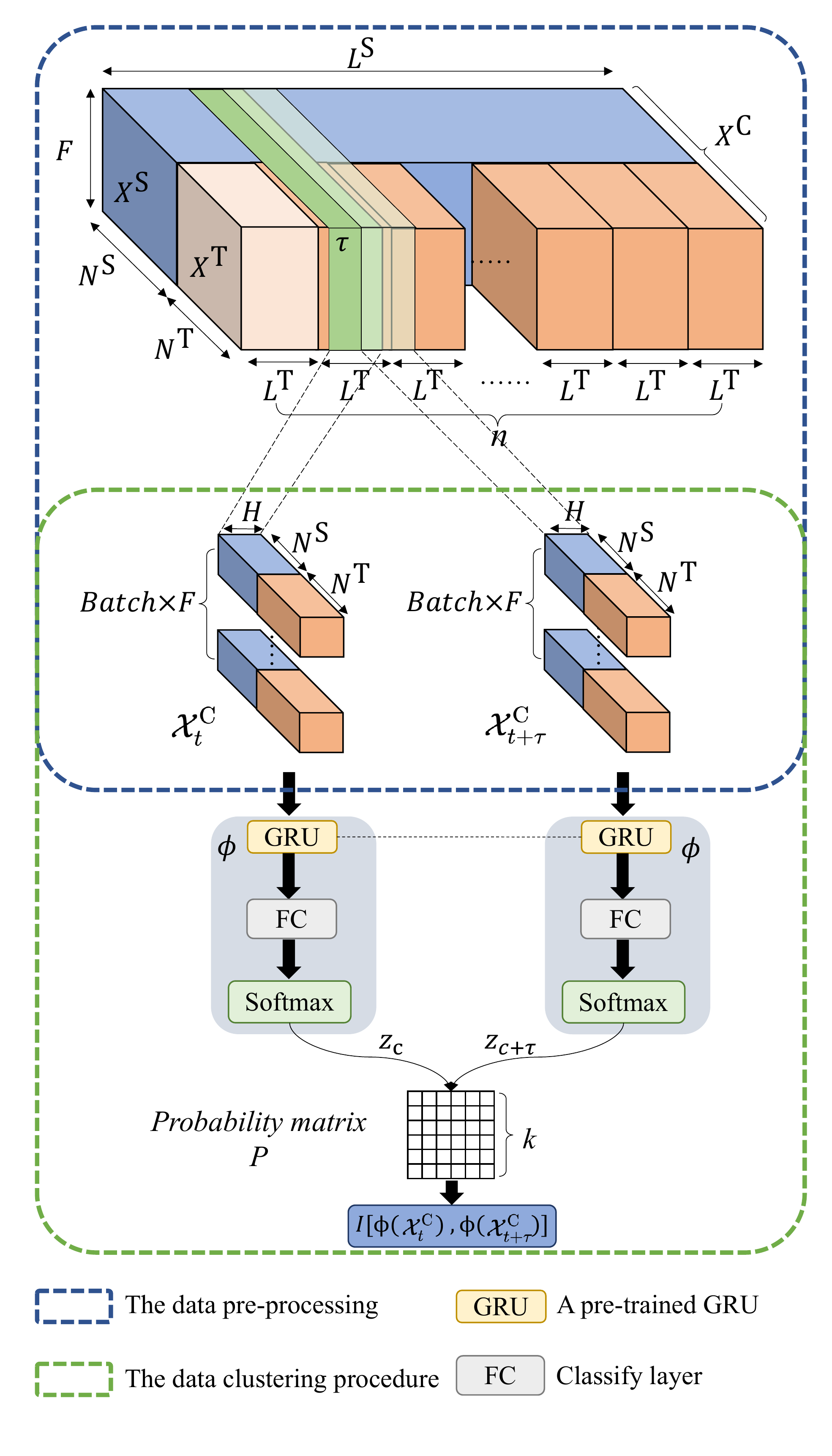}
	\caption{Data pre-processing and node aggregation in TCB.}
	\label{fig:TCBinp}
  \end{figure}

With regard to the first challenge (data distribution difference), we propose Temporal Cluster Block (TCB) to cluster all potential shared factors based on a hidden feature generated by a pre-trained GRU as shown in Fig. \ref{fig:TCBinp}. We next add a layer FC as a classification layer, which will be retrained with a new number of clusters. The training optimization objective is to maximize the mutual information between the groups obtained after the left and right parts GRU-FC-Softmax models. A comprehensive explanation of the method will be displayed in the following paragraphs. In addition, the process of clustering nodes is referred to as \textit{Node Aggregation} in the sequel.

It is worth noting that the traffic data are naturally unlabeled, limiting the possibility of aggregating nodes in the source and target cities with similar trends only based on the feature data. 
Therefore, the problem becomes how to use unsupervised learning to cluster similar nodes belonging to two different distributed datasets. 
To that end, we propose creating a comprehensive dataset that includes both source and target data to eliminate the distribution gap. 
The samples from the comprehensive dataset contain all nodes of the source and target cities. 
Following that, we employ a contrastive learning approach to classify all nodes in this comprehensive dataset.
Contrastive learning is an unsupervised learning method based on mutual information. 
The basic idea is to cross-reflect the object's hidden features using different forms of the same object. See refs \cite{GCLempirical,GCLrepresentation,GCLaugmentation} for more detailed introduction.

During the process of node aggregation, we keep the set of classification results with the largest mutual information and regard it as inputs for the following blocks. In the sequel, we first give the original definition of mutual information. Then we describe the data pre-processing before node aggregation and, finally, the mutual information computation in this process.
Fig. \ref{fig:TCBinp} illustrates the overall framework of TCB.

Mutual information refers to the infomax principle proposed by Linsker \cite{Linsker1988SelforganizationIA}, which plays a vital role in the quality of the representations constructed and learned by generative models \cite{Hjelm2019LearningDR}. 
%
The detailed definition of mutual information is as follows:
\begin{definition}[Mutual information]
	For two discrete variables $X$ and $Y$ whose joint probability distribution is $\mathbb{P}_{XY}(x,y)$, the mutual information between them, denoted by $I(X;Y)$, is defined by
	\begin{equation}
		I(X;Y) = \sum_{x,y}\mathbb{P}_{XY}(x,y)log \frac{\mathbb{P}_{XY}(x,y)}{\mathbb{P}_X(x,y)\mathbb{P}_Y(x,y)}.
	\end{equation}
\end{definition}
Mutual information quantifies the probability that these two variables share information. In the TCB module, we use it as the loss function for the clustering network $\Phi$ to select the best number of clusters $k$ and calculate each node belonging cluster.

\subsubsection{Data Pre-processing}
Before aggregating nodes from multiple cities, we pre-process and align the raw data for easier subsequent computation.
Without any data pre-processing, traffic speed data of source and target cities are shaped as shown in Fig. \ref{fig:transferdata}. We assume the source data to be $n$ times the length of the target data along the time dimension, i.e., $n$ is the ratio of source-target data quantities. To make them of the same lengths, we employ a na\"ive resampling method and replicate $(n-1) \times X^{\mathcal{T}}$ along the time dimension to increase the length, i.e., $n \cdot L^{\mathcal{T}} = L^{\mathcal{S}}$ as shown in Fig. \ref{fig:TCBinp}. 
(The light orange block is copied to dark orange ones and stacked to reach the same length as the blue one, where $n$ is the number of orange blocks.)
Subsequently, we concatenate the source and target city data (the blue and orange blocks) to make $X^{\textnormal{C}} = [X^{\mathcal{S}}, X^{\mathcal{T}}_1\Vert{}X^{\mathcal{T}}_2\Vert{}\dots\Vert{}X^{\mathcal{T}}_n] \in \mathbb{R}^{L^{\mathcal{S}} \times (N^{\mathcal{S}} + N^{\mathcal{T}}) \times F}$, where $X^{\mathcal{T}}_i$ is the $i$-th replication of the original $X^{\mathcal{T}}$.
The new raw dataset $X^{\textnormal{C}}$ contains a total of $N^{\mathcal{S}}+N^{\mathcal{T}}$ nodes.
Following the idea of contrastive learning that applies different forms (views) of the same object to cross-reflect the object's hidden features, we sample two subsets from the comprehensive dataset $X^{\textnormal{C}}$ with time difference $\tau$, where $0 < \tau < H$. 
In traffic prediction, we predict the traffic states at future times with historical data containing latent information about the future data. Whereas the two subsets with a time gap of $\tau$ have overlapping parts, i.e., they share potential future information. Consequently, their predicted classification results should be intuitively similar.
In this work, the comprehensive dataset $X^{\textnormal{C}}$ is an object whose two subsets with gap $\tau$ are different forms of itself. We expect the classification probabilities of each node learned from these two subsets to be as consistent as possible and to maintain maximum mutual information.

We define these two subsets which sample from the comprehensive dataset as follows:
The immediate past $H$ historical observations at time step $t$ is the first subset $\mathcal{X}^{\textnormal{C}}_{t} = (x^{\textnormal{C}}_{t-H}, \cdots, x^{\textnormal{C}}_{t-2}, x^{\textnormal{C}}_{t-1}) \in \mathbb{R}^{H \times (N^{\mathcal{S}} + N^{\mathcal{T}}) \times F}$.
Take another $H$ historical observation traffic states at time step $t+\tau$ as the second subset $\mathcal{X}^{\textnormal{C}}_{t+\tau} = (x^{\textnormal{C}}_{t+\tau-H}, \cdots, x^{\textnormal{C}}_{t+\tau-2}, x^{\textnormal{C}}_{t+\tau-1}) \in \mathbb{R}^{H \times (N^{\mathcal{S}} + N^{\mathcal{T}}) \times F}$ after $\tau$.
Then, we combine the two subsets $\mathcal{X}^{\textnormal{C}}_{t}$ and $\mathcal{X}^{\textnormal{C}}_{t+\tau}$ into a pair of input data $(\mathcal{X}^{\textnormal{C}}_{t}, \mathcal{X}^{\textnormal{C}}_{t+\tau})$.

\subsubsection{Node Aggregation}

$(\mathcal{X}^{\textnormal{C}}_{t}, \mathcal{X}^{\textnormal{C}}_{t+\tau}) \in X^{\textnormal{C}}$ is a paired data sampled from a joint probability distribution $P(\mathcal{X}^{\textnormal{C}}_{t}, \mathcal{X}^{\textnormal{C}}_{t+\tau})$. 
We devise a neural network called Cluster Net $\Phi$ to extract features from the pair of data, making probabilistic predictions about classifying a couple of data with the same distribution and reinforcing the mutual information collected in the classification results.
The network comprises a pre-trained GRU, a Fully Connected (FC) layer as a classification layer, and a softmax activation layer. 
Specifically, GRU aims to extract hidden temporal information. FC and the softmax active layers generate the probability of which clusters the nodes belong to. FC layer is retrained every time with a different number of clusters $c$.
We further acquire probabilistic information of classifications when the inputs are $\mathcal{X}^{\textnormal{C}}_t$ and $\mathcal{X}^{\textnormal{C}}_{t+\tau}$, respectively by a cluster network $\Phi$.
Following that, we calculate the likelihood (mutual information) of classification possibility obtained by network $\Phi$, namely, $\mathcal{I}[\Phi(\mathcal{X}^{\textnormal{C}}_t, \mathcal{X}^{\textnormal{C}}_{t+\tau})]]$ depicted in Fig. \ref{fig:TCBinp}.
The output classification probabilities $\Phi_{c}(\mathcal{X}) \in [0, 1]^{c}$ over $c$ clusters can be represented by a distribution of a discrete random variable $z$, namely, $\mathbb{P}(z=c |\mathcal{X}) = \Phi_{c}(\mathcal{X})$, where $c$ is determined by mutual information, and $z$ is the representative feature of the input data.
For $\mathcal{X}^{\textnormal{C}}_t$ and $\mathcal{X}^{\textnormal{C}}_{t+\tau}$, $z_{\textnormal{c}}$ and $z_{\textnormal{c}+\tau}$ are their discrete representations, respectively.
With the pair of input data $(\mathcal{X}^{\textnormal{C}}_t, \mathcal{X}^{\textnormal{C}}_{t+\tau})$, the conditional joint distributions $\mathbb{P}(z_{\textnormal{c}} =c, z_{\textnormal{c}+\tau} =c' | \mathcal{X}^{\textnormal{C}}_t, \mathcal{X}^{\textnormal{C}}_{t+\tau}) = \Phi_c(\mathcal{X}^{\textnormal{C}}_t)\cdot \Phi_c(\mathcal{X}^{\textnormal{C}}_{t+\tau})$ are specified. 
This formula suggests that $z_{\textnormal{c}}$ and $z_{\textnormal{c}+\tau}$ should be independent. 
However, for the network $\Phi$, the inputs $\mathcal{X}^{\textnormal{C}}_t$ and $\mathcal{X}^{\textnormal{C}}_{t+\tau}$ obey the same distribution since both of them are taken from the dataset $X^{\textnormal{C}}$. Hence, $z_{\textnormal{c}}$ and $z_{\textnormal{c}+\tau}$ are not entirely independent.
This accords with the intuition that we can learn from historical data to predict future data since both contain common latent features. Accordingly, the classification probability obtained by the same network $\Phi$ should be similar when the two historical data are developed with a time difference $\tau$. In other words, the predicted classification probability for $z_{\textnormal{c}}$ can alternatively be indicated by $z_{\textnormal{c}+\tau}$.
The joint probability distribution $\mathbb{P}(z_{\textnormal{c}} =c, z_{\textnormal{c}+\tau}=c')$ is given by the $c \times c$ matrix $\bm{P}_{cc'}$,
\begin{align}
    \bm{P}_{cc'} = \frac{1}{m}\sum_{i=1}^{m}\Phi_{c}(\mathcal{X}^{\textnormal{C}}_t)\cdot \Phi_{c}(\mathcal{X}^{\textnormal{C}}_{t+\tau})^{\top},
\end{align}
where $m$ is the number of samples.
Summing over the rows and columns of this matrix yields marginal probabilities $\bm{P}_{c} = \mathbb{P}(z_{\textnormal{c}} = c)$ and $ \bm{P}_{c'} = \mathbb{P}(z_{\textnormal{c}+\tau} = c') $. For convenient calculation, $\bm{P}_{cc'}$ is symmetrized by
\begin{align}
    \hat{\bm{P}}_{cc', jk} = \hat{\bm{P}}_{cc', kj}= \frac{\bm{P}_{cc', jk} + \bm{P}_{cc', kj}}{2},
\end{align}
where $j, k$ are the indices of rows and columns of $\bm{P}_{cc'}$.
The objective function can be correspondingly derived as follows:
\begin{align}
\label{eq:mi}
    \mathcal{I}(z_{\textnormal{c}},z_{\textnormal{c}+\tau}) = \mathcal{I}(\hat{\bm{P}}_{cc'}) = \sum_{1}^{c}\sum_{1}^{c'}\hat{\bm{P}}_{cc'}\ln{\frac{\hat{\bm{P}}_{cc'}}{\bm{P}_{c} \cdot \bm{P}_{c'}}},
\end{align}

\renewcommand{\algorithmicrequire}{\textbf{Input:}} 
\renewcommand{\algorithmicensure}{\textbf{Output:}} 
\begin{algorithm}[!t]
	\caption{Temporal Cluster Block (TCB)}
	\label{alg:tcb}
	\begin{algorithmic}[1]
		\REQUIRE ~~\\{Data $\mathcal{X}^{\textnormal{C}}_{t}$, $\mathcal{X}^{\textnormal{C}}_{t+\tau}$, \\
		the number of clusters $c \in \left\{c_{1}, c_{2}, \ldots,  c_{m}\right\}$}	
		\ENSURE ~~\\{Clustering result $\mathcal{C}=\left\{C_{1}, C_{2}, \ldots C_{k}\right\}$.\\
		the optimal number of cluster $c$}
		\STATE Initialize $\mathcal{I}_{\textnormal{bst}} = \text{None}$, $\mathcal{C} = \text{None}$ and $c_{bst}= \text{None}$
		\FOR{each $c$} 
		\REPEAT
	    \STATE Calculate the mutual information $\mathcal{I}$ 
	    \IF{$\mathcal{I}_{\textnormal{bst}}$ is $None$ $\OR$ $ \mathcal{I}_{\textnormal{bst}} < \mathcal{I}$}
	    \STATE $\mathcal{C}_{\textnormal{bst}} \gets \mathcal{C}$
	    \STATE $c_{bst} \gets c$
	    \ENDIF
	    \UNTIL {convergence}
		\ENDFOR
		
		\RETURN $c_{bst} \text{ and } \mathcal{C}_{bst}$.
	\end{algorithmic}
\end{algorithm}

The goal of TCB is to maximize \eqref{eq:mi} for the best clustering results and the number of clusters with the largest mutual information.
The algorithm is shown in Alg. \ref{alg:tcb}.

\subsection{Graph Reconstruction Block}\label{sec:GRB}
With respect to the second challenge (prior knowledge error), we propose a new scheme for cluster correction on the distance to compensate for the error. Besides, the scheme also corrects accumulated errors in TCB.
In Sec \ref*{sec:TCB}, we presented employing mutual information for node aggregation of multiple cities' mixed data.
The clustering relies entirely on validating the mutual information without using the cluster label of the raw data. 
While it is true that no label exists in practice to verify the clustering results, the minor clustering error may accumulate and lead to notable performance degradation if not properly counteracted.
This issue needs to be resolved for better prediction transferability, and we rely on the proposed GRB to address the two issues.


Distance representation is commonly adopted in various clustering and classification tasks. 
Among the representations, Euclidean and Manhattan distances are more successful than others for Euclidean-structured data like images. A variety of improved versions have been developed to aid image processing.
However, time-series data is not naturally fit for measuring Euclidean distance. 
This is because time-series are more susceptible to temporal transformations, such as shifts and practical scaling than others, which are inevitable during data collection. 
If the error between data is established based on Euclidean distance, it may not accurately reflect their similarity. 
In contrast, DTW is more frequently used to determine the time-series distance.

In the previous section, we employed $H$ historical traffic states for learning data periodicity. For most studies, $H$ historical traffic states tend only to contain information for a limited duration, e.g., 1 hour.
On the other hand, the periodicity is statistically observed in other time spans, e.g., days, weeks, and even years.
As a result, we apply the DTW approach in GRB to reconstruct the graph structure among nodes based on long-term internode correlation and strengthen the connection of their periodical similarities.

For cluster $C_{i}, i \in [1,2,\cdots,c]$ that contains $N_{i}^{\textnormal{C}} = N_{i}^{\mathcal{S}} + N_{i}^{\mathcal{T}}$ nodes (includes source and target), GRB first randomly selects one day of data on each node. Let $D$ be the length of one day's data\footnote{We use one day instead of weeks or years due to heavy computation and scarce target city data concerns.}, and $D \leqslant L^{\mathcal{T}}$.
%
For any two nodes $N_{\textnormal{1}}, N_{\textnormal{2}} \in \mathbb{R}^{D \times 1}$ in $C_{i}, i \in [1,2,\cdots,c]$, we compute their distance matrix $\bm{M} \in \mathbb{R}^{D \times D}$, $\bm{M}_{jk} = \left| N_{\textnormal{1},j} - N_{\textnormal{2},k} \right|, j,k \in [1,2,\cdots,D]$. $N_{\textnormal{1},j}$ is the $j$-th speed of node $N_{\textnormal{1}}$ and $N_{\textnormal{2},k}$ is the $k$-th speed of node $N_{\textnormal{2}}$.
The next step is to find the shortest path from $\bm{M}_{11}$ to $\bm{M}_{DD}$ as follows:
\begin{equation}
    \begin{aligned}
    L_{\textnormal{min}}(1, 1) = &\bm{M}_{11}; \\
    L_{\textnormal{min}}(j, k) = &\min\{L_{\textnormal{min}}(j-1, k),  L_{\textnormal{min}}(j, k-1), \\ &L_{\textnormal{min}}(j-1, k-1)\}  + \bm{M}_{jk},
    \end{aligned}
\end{equation}
where $L_{\textnormal{min}}(j, k)$ represents the minimized distance from $\bm{M}_{11}$ to $\bm{M}_{jk}$, and $L_{\textnormal{min}}(D, D)$ is the minimum distance between nodes $N_{\textnormal{1}}$ and $N_{\textnormal{2}}$.

The shortest distance $L_{\textnormal{min}}(D, D)$ of every two nodes in the cluster $C_{i}$ is calculated to form a similarity matrix $\bm{W}=\{\bm{W}_{mn}\} \in \mathbb{R}^{N^{\textnormal{C}}_i \times N^{\textnormal{C}}_i}$, where $\bm{W}_{mn}$ represents the minimum distance between node $N_m \text{ and } N_n, m,n \in [1,2,\cdots, N^{\textnormal{C}}_i]$.
The adjacency matrix $\bm{A}$ can be obtained by distance calculation from geographic location information and can also be regarded as a distance similarity matrix.
In TrafficTL, to attenuate the adverse influence of incorrect prior adjacency matrix and to bolster the temporal trend connectedness between nodes, we use the similarity matrix $\bm{W}$ in place of $\bm{A}$ for convolution operation on graphs.

\subsection{Ensemble Model} \label{sec:ensemble}
\begin{figure}[!t]
	\centering
	\includegraphics[scale=0.38]{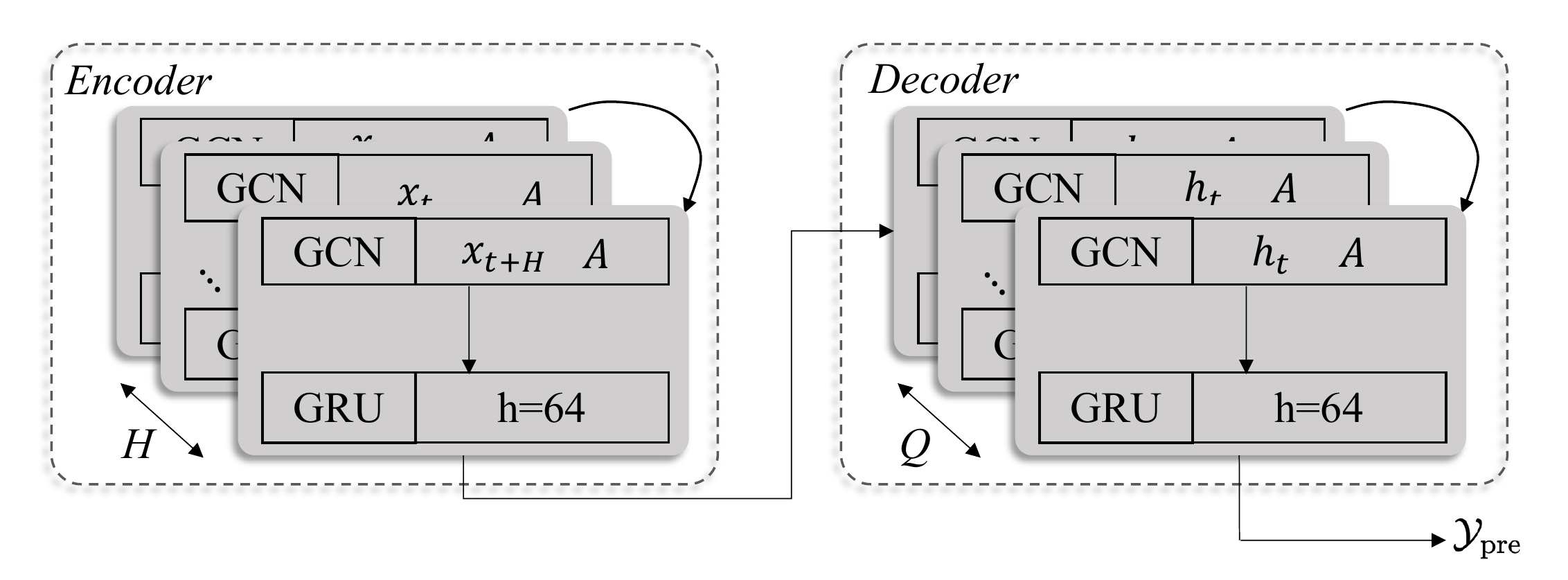}
	\caption{The architecture of network $\Psi$'s and $\Psi'$'s in ensemble models. $H$ historical traffic states with their adjacency matrix $\bm{A}$ are fed into the connective GCN-GRU layers. $h$ is the hidden size of the GRU layer. The final output of encoder $h_t$ contains the past $H$ temporal and spatial information. And we get the prediction $\mathcal{Y}_{\textnormal{pre}}$ through the decoder to decode the hidden state $h_t$.}
	\label{fig:ensemblebackbone}
  \end{figure}

In response to the third challenge (indistinguishable parameters), we propose an ensemble scheme to minimize the differences between data of multiple cities by training various models with different clustered subsets to reduce the negative impact of indistinguishable shared parameters in one model.
Each cluster in this design (depicted in Fig. \ref{fig:overview}) has its network $\Psi$ for training and sharing parameters with the transferable network $\Psi'$. The whole complete model is stacked by $c$ $\Psi$s and $\Psi'$s.
The architecture of networks $\Psi$ and $\Psi'$ in TrafficTL is shown in Fig. \ref{fig:ensemblebackbone}.
It is composed of an encoder and a decoder.
The encoder includes $H$ GCN-GRU connection layers, each receiving an observation state $x_t, t \in (1,2,\cdots,H)$, and then saves and delivers the encoded information to the following layer. 
The decoder comprises $Q$ GCN-GRU layers to recover encoded information and convert it to the following $Q$-steps traffic states.
Among the layers, GCN is a type of neural network that captures spatial information on topological graph structures. 
The spatial information of the topological graph is represented by the adjacency matrix $\bm{A}$, which records the node connections.
Additionally, GRU is a variant structure of LSTM and RNN which alleviates the gradient disappearance and explosion problem and has better results in solving the long-term dependence problem over canonical RNNs.

\subsubsection{GCN layer} \label{sec:GCN}

In GCN layers, we denote $\bm{A}$ as the adjacency matrix, and $\tilde{\bm{A}} = \bm{A} + \bm{I}_{\textnormal{N}}$ is the adjacency matrix with added self-connections, where $\bm{I}_{\textnormal{N}}$ is the identity matrix. Thereby, the graph Laplacian matrix can be computed as:
\begin{align}
	\bm{L} = \bm{D}^{-\frac{1}{2}} \tilde{A} \bm{D}^{-\frac{1}{2}},
\end{align}
where $\bm{D} = diag(\sum_j \bm{A}_{ij})$ is the diagonal degree matrix.

For each graph convolution operation, we have
\begin{align}\label{gcn}
    H^{(l+1)} = \sigma(\bm{L}H^{(l)}\theta^{(l)}),
\end{align}
where $l$ is the layer index, $H^{(l)}$ stands for the output of the $l$-th layers, $\theta^{(l)}$ represents the $l$-th layer's parameters, and $\sigma(\cdot)$ is the sigmoid function. Combined with \eqref{gcn}, the output is calculated as follows:
\begin{align}
    f(\bm{X}_t, \bm{A}) = \sigma(\bm{L} \operatorname{ReLU}(\bm{L} X W_0)W_1),
\end{align}
where $f(\bm{X}_t,\bm{A})$ is the output, $\bm{X}_t$ is the feature matrix of the set $\mathcal{X}^{\mathcal{S}_i}_t \text{ and } \mathcal{X}^{\mathcal{T}_i}_t$, $W_0$ stands for the weight matrix mapping input to the hidden unit, and $W_1$ represents the weight of the next layer. $\operatorname{ReLU}(\cdot)$ is the Rectified Linear Unit \cite{relu}.
%
It is worthwhile to note that the involved adjacency matrix $\bm{A}$ is reconstructed as previously introduced in Sec \ref*{sec:GRB} considering the lack of target city data.

\subsubsection{GRU layer}
GRU contains a reset gate $r$ and an update gate $z$. The former aims to combine new information with previous information, and the latter aims to choose the information that needs to be remembered. The two gates can be expressed as
\begin{align}
	r_{t} = \sigma (W_r \cdot [h_{t-1}, \bm{X}_{t}]),\\
	z_{t} = \sigma (W_z \cdot [h_{t-1}, \bm{X}_{t}]),
\end{align}
where $\sigma$ represents the sigmoid activate function, $h_{t-1}$ is the previous state at time step $t-1$, and $\bm{X}_{t}$ is the input time-series feature matrix at time $t$. Through the reset gate $r_{t}$, we can get the next state $ \hat{h}_{t} $ which contains part of previous information determined by $r_{t}$
\begin{align}
	\hat{h}_{t} = \tanh([r_{t} \odot h_{t-1}, \bm{X}_{t}]),
\end{align}
where $\odot$ represents the Hadamard product, the $\tanh$ function can alleviate the gradient explosion and reduce the computation. The update state at the current step is as follows:
\begin{align}
	h_t = (1 - z_{t}) \odot h_{t-1} + z_{t} \odot \hat{h}_{t}.
\end{align}

\subsubsection{Transfer process}
As shown in Fig. \ref{fig:overview}, we take nodes belonging to the source city in cluster $C_i$ and their adjacency matrix $\bm{A}^{\mathcal{S}_i}$ to make up the input data $\mathcal{X}^{\mathcal{S}_i}_t$ for training network $\Psi_i$. $\mathcal{Y}^{\mathcal{S}_i}_{\textnormal{pre}}$ is a set of the following $Q$ traffic states' predictions at time $t$. Network $\Psi_i$ is optimized by minimizing the $L_1$ loss between $\mathcal{Y}^{\mathcal{S}_i}_{\textnormal{pre}}$ and $\mathcal{Y}^{\mathcal{S}_i}_{\textnormal{true}}$, where $\mathcal{Y}^{\mathcal{S}_i}_{\textnormal{true}}$ is the $Q$ traffic states' ground truth.
%
Subsequently, we replace a portion (defined by a control parameter $\beta$) of the source city nodes $\mathcal{X}^{\mathcal{S}_i}_t$ with higher similarity to the target city nodes (or randomly replace some nodes in the source city), shown as red dots in Fig. \ref{fig:overview}.
The predictive network $\Psi_i$ shares parameters with the transferable network $\Psi'_i$.  $\mathcal{Y}^{\mathcal{S}_i+\mathcal{T}_i}_{\textnormal{pre}}$ is the prediction of the mix-cities input data. Evaluation metrics are calculated by the forecasting results of the target city $\mathcal{Y}^{\mathcal{T}_i}_{\textnormal{pre}}$ with their ground truth $\mathcal{Y}^{\mathcal{T}_i}_{\textnormal{true}}$. The whole process can be described as follows:
\begin{align}
\mathcal{X}^{\mathcal{S}_i+\mathcal{T}_i}_{t} \leftarrow & \operatorname{Rep}(\beta\cdot\mathcal{X}^{\mathcal{S}_i}_{t}),\\
\Psi_i \stackrel{\theta_i}{\leftrightarrow} & \Psi'_i,\\
\mathcal{Y}^{\mathcal{T}_i}_{\textnormal{pre}} = & \Psi'_i(\mathcal{X}^{\mathcal{S}_i+\mathcal{T}_i}_{t}),\\
\mathcal{M}_i = & \operatorname{Evaluation}(\mathcal{Y}^{\mathcal{T}_i}_{\textnormal{pre}},\mathcal{Y}^{\mathcal{T}_i}_{\textnormal{true}}),
\end{align}
where $\operatorname{Rep}(\cdot)$ means replacing $\beta$ of $\mathcal{X}^{\mathcal{S}_i}_{t}$ with target nodes, $\theta_i$ is the shared parameters in predictive network $\Psi_i$ and transferable network $\Psi'_i$, $\mathcal{M}_i$ is a set of metrics of cluster $C_i$ and $\operatorname{Evaluation}(\cdot, \cdot)$ stands for the performance evaluation metric.
At last, we summarize the overall metric $\mathcal{M}$ by,
\begin{align}
    \mathcal{M}=\frac{1}{c} \sum_{i=1}^{c} \mathcal{M}_{{i}},
\end{align}
where $c$ is the number of clusters.
We also provide the algorithm of Ensemble models in Alg. \ref{alg:enseble}.

\renewcommand{\algorithmicrequire}{\textbf{Input:}} 
\renewcommand{\algorithmicensure}{\textbf{Output:}} 
\begin{algorithm}[!t]
	\caption{Ensemble models algorithm}\label{alg:enseble}
	\begin{algorithmic}[1]
		\REQUIRE ~~\\{Clustering set $\mathcal{C}=\left\{C_{1}, C_{2}, \ldots C_{\textnormal{c}}\right\}$}
		\ENSURE ~~\\{metrics $\mathcal{E}$}
		\STATE Initialize sub-model parameters $\theta_{c}$
		\FOR{each $C \in \mathcal{C}$} 
		\REPEAT
		\STATE Train model with data $\mathcal{X}^{\mathcal{S}} \in C$
	    \STATE Calculate the loss
	    \STATE Update model parameters $\theta_k$
	    \UNTIL {convergence}
	    \STATE Save sub-model $M_{c}$
	    \ENDFOR
	    
	    \FOR{each $C \in \mathcal{C}$}
	    \FOR{round $1,2,\ldots$}
	    \STATE Replace $\beta$ of all nodes in $\mathcal{X}^{\mathcal{S}}$ with $\mathcal{X}^{\mathcal{T}}$
	    \STATE Get prediction $\mathcal{Y}^{\mathcal{S}+\mathcal{T}}_{\textnormal{pre}}$ with sub-model $M_{c}$
	    \STATE Calculate metric between $(\mathcal{Y}^{\mathcal{T}}_{\textnormal{pre}}, \mathcal{Y}^{\mathcal{T}}_{\textnormal{true}})$
	    \ENDFOR
	    \STATE Calculate average cluster metric $\mathcal{M}_{i}$
	    \ENDFOR
	    \STATE Calculate average total cluster metric $\mathcal{M}=\frac{1}{c} \sum_{i=1}^{c} \mathcal{M}_{{i}}$.
		\RETURN $\mathcal{M}$
	\end{algorithmic}
\end{algorithm}

\section{Case Studies}\label{section:V}
This section compares the proposed TrafficTL with current state-of-the-art over three widely-recognized datasets. Next, we evaluate the effectiveness of TCB and GRB modules. Finally, we investigate the model sensitivity to hyper-parameters.

\subsection{Dataset and Configurations}\label{sec:setup}
In the case studies, we adopt three real-world traffic speed datasets as follows.
\begin{itemize}
    \item \textbf{Nav-BJ:} The dataset collects the average vehicle speeds of $1,362$ on major roads in Beijing, China, from January 1st to July 1st, 2019. Specifically, there are two sets of available data, the adjacency matrix and the feature matrix. The former is binary and consists of $1,362$ roads. The $52,128 \times 1,362$ feature matrix contains $8,990$ time steps, and row data exhibits the speed of all roads at the corresponding time. 
    \item \textbf{Nav-SH:} The dataset includes the average vehicle speeds of $1,401$ on major roads in Shanghai, China, from January 1st to May 1st, 2019. The feature matrix's shape in Nav-SH is $34,560 \times 1,401$, and the adjacency matrix shape is  $1,401 \times 1,401$. We employ the feature matrix as training data. 
    \item \textbf{Nav-HZ:} The dataset includes the average speed in Hangzhou from January 1st to July 1st, 2019. The survey data includes $413$ major roads in Hangzhou, China. It includes a $ 52,128 \times 413$  feature matrix and a $413 \times 413$ adjacency matrix.
\end{itemize}

\begin{figure}[h!]
  \centering
  \includegraphics[width=1.0\linewidth]{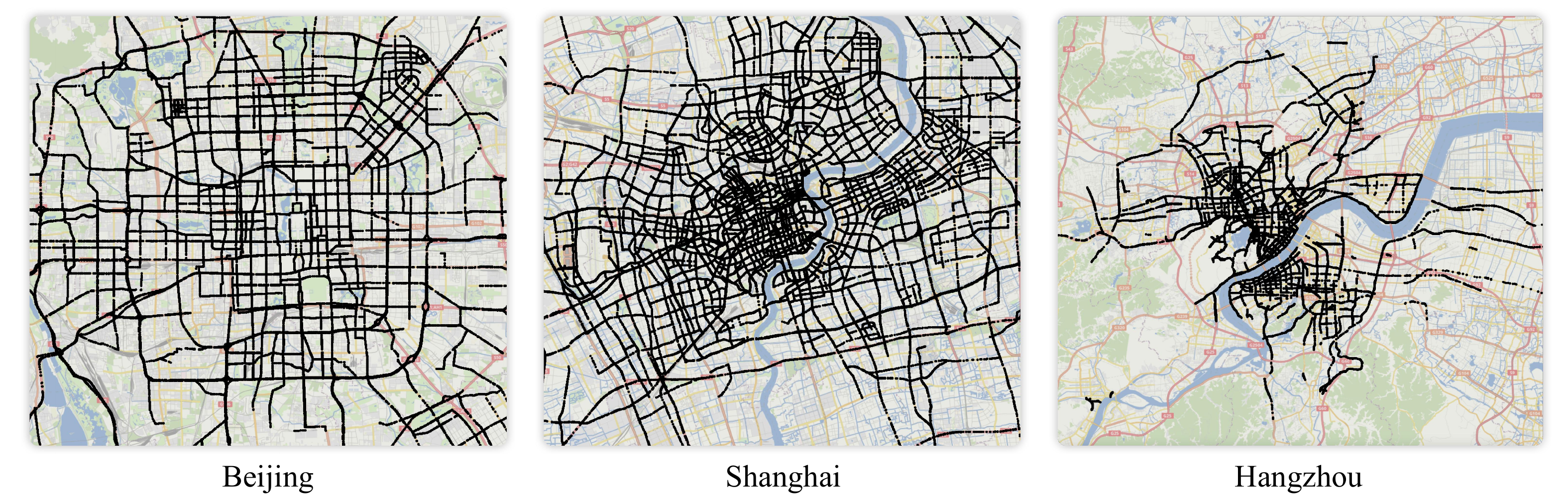}
  \caption{Overview of the three investigated cities with road segments colored by black.}
  \label{fig:roadmap}
\end{figure}

All three datasets were collected with a constant \SI{5}{\minute} interval. 
The main collection roads of the three datasets are visualized in Fig. \ref{fig:roadmap}. We investigate the effect of transfer learning on traffic prediction for cities with scarce data. 
In particular, we use Nav-SH, which contains fewer data in time horizon, and Nav-HZ, with fewer roads ($413$) as the target cities to be transferred to.
We use only one day of data (i.e., $288$ time steps) from the target city as the validation dataset and ten days of data as the test dataset. 
In order to obtain the correlation on the time trend, max-min normalization is applied to the data.

All experiments are conducted on a Linux server with Intel E5-2620v4 CPUs and GeForce RTX 2080Ti GPUs. All baselines and the proposed model are built with Pytorch 1.7.0 and Python 3.8.3.
When making predictions, we use historical data of the immediate past hour to forecast the next \SI{15}{\minute}/\SI{30}{\minute}/\SI{60}{\minute} traffic speed, namely, $H = 12$, $Q = 3/6/12$, respectively. We train our model with the Adam optimizer. The initial learning rate is set to $0.05$, the batch size is $64$, the maximum training epoch for proposed models is $50$, and the early stopping strategy is adopted where the training is terminated if the best result is not updated for consecutive $5$ epochs. The cluster number is $10$. 
The percentage $\beta$ of the replaced nodes is $0.2$.

\subsection{Evaluation Metrics}
\label{sec:metrics}
We adopt the following three metrics to evaluate the performance of models:
\begin{enumerate}
    \item Mean Absolute Error (MAE):
    \begin{equation}
    \text{MAE} = \frac{1}{m} \sum^{m}_{i=1}\left|(Y_{i} - \hat{Y}_{i})\right|
    \end{equation}
    \item Root Mean Squared Error (RMSE):
    \begin{equation}
    \text{RMSE} = \sqrt{\frac{1}{m} \sum^{m}_{i=1}(Y_{i} - \hat{Y}_{i})^2 }
    \end{equation}
    \item Mean Absolute Percentage Error (MAPE):
    \begin{equation}
        \text{MAPE} = \frac{100\%}{m} \sum_{i=1}^{m} \left| \frac{\hat{Y}_{i}-Y_{i}}{Y_{i}}\right|
    \end{equation}
\end{enumerate}
where $Y_{i}$ and $\hat{Y}_{i}$ are ground truth and predicted speed, and $m$ is the number of samples.

\subsection{Baseline Models}
We compare the proposed model with two categories of methods: non-transfer learning and transfer learning methods. For non-transfer learning methods, we employ T-GCN, ASTGCN, DCRNN and XGBoost as baselines. For transfer learning models, we choose RegionTrans, MetaST, and TL-DCRNN:
\begin{itemize}
    \item T-GCN\cite{Zhao2020TGCNAT}: A temporal graph convolutional network model combining GCN with GRU.
    \item ASTGCN\cite{ASTGCN}: A Spatial-Temporal Graph Convolution Network model based on a self-attention mechanism.
    \item DCRNN\cite{Li2018DiffusionCR}: A traffic predicting model capturing the spatial dependency with bidirectional random walks on the graph.
    \item XGBoost\cite{xgboost16}: An efficient gradient-boosting decision tree algorithm that uses multiple trees to make decisions together.
    \item RegionTrans\cite{Wang2019CrossCityTL}: A transferable model with knowledge transfer that matches source cities with similar regions in target cities by learning city-region matching functions.
    \item MetaST\cite{Yao2019LearningFM}: A spatial-temporal network with a meta-learning paradigm learns the generic initialization of a spatio-temporal network to adapt to the target city and achieve knowledge transfer.
    \item TL\_DCRNN\cite{Mallick2021TransferLW}: A model for transferable traffic prediction based on DCRNN model, and transfer the knowledge between source and target cities with graph partitioning and clustering.
    
\end{itemize}

\begin{table*}[!t]
\centering
\caption{The Performance of TrafficTL and Baseline Approaches}
\label{tbl:perform}
\centering
\begin{tabular}{clccccccccc}
\toprule
& & \multicolumn{9}{c}{Nav-BJ to Nav-SH}  \\                               \cmidrule(lr){3-11}   
& & \multicolumn{3}{c}{\SI{15}{\minute}}  & \multicolumn{3}{c}{\SI{30}{\minute}} & \multicolumn{3}{c}{\SI{60}{\minute}}\\
\cmidrule(lr){3-5}\cmidrule(lr){6-8}\cmidrule(lr){9-11}
& &  MAE & RMSE & MAPE  & MAE & RMSE & MAPE & MAE & RMSE & MAPE  \\ \cmidrule(lr){1-11}
\multicolumn{1}{c}{\multirow{6}{*}{Non-transfer}} & TGCN     & 6.46   & 8.69  & \multicolumn{1}{c}{23.49} & 6.48  & 8.73  & \multicolumn{1}{c}{23.52} & 6.56  & 8.78  & 24.02 \\
\multicolumn{1}{c}{}  & ASTGCN   & 6.48  & 8.69  & \multicolumn{1}{c}{23.74} & 6.49   & 8.72  & \multicolumn{1}{c}{23.65} & 6.57  & 8.78  & 24.16 \\
\multicolumn{1}{c}{} & DCRNN    & \uline{3.79}  & 6.20  & \multicolumn{1}{c}{15.67} & \uline{4.00}  & 6.33  & \multicolumn{1}{c}{16.25} & 4.76  & 6.96  & 19.16 \\
\multicolumn{1}{c}{} & TGCN$^\dagger$     & 7.34  & 9.86  & \multicolumn{1}{c}{26.78} & 7.56 & 9.98 & \multicolumn{1}{c}{27.86}  & 8.01  & 10.45 & 28.96 \\
\multicolumn{1}{c}{}  & ASTGCN$^\dagger$   & 7.86  & 10.23 & \multicolumn{1}{c}{28.78} & 7.94  & 10.37 & \multicolumn{1}{c}{29.15} & 8.23  & 10.43 & 29.55 \\ 
\multicolumn{1}{c}{} & DCRNN$^\dagger$    & 4.38  & 6.54  & \multicolumn{1}{c}{17.12} & 6.23  & 8.02  & \multicolumn{1}{c}{18.84} & 6.53  & 8.64  & 20.13 \\ 
& XGBoost  & 4.33  & 6.51  & 14.53 & 4.45  & 6.63  & 14.93 & 4.59  & 6.78  & 15.35 \\ 
\cmidrule(lr){1-11}                                    
\multicolumn{1}{l}{\multirow{4}{*}{Transfer}}    & TL\_DCRNN &  6.90 & 9.76 & 22.92 & 7.61 & 10.85  & 25.21 & 7.90 & 10.76 & 28.21  \\
& MetaST & 4.29 & 6.17  & 13.20 & 4.37  & 6.27  & 13.47 & 4.48  & 6.40  & 13.86  \\ 
& RegionTrans & 4.24 & 6.11  & 13.34 & 4.79  & 6.71  & 13.64 & 5.07  & 7.18  & 14.03  \\ 
\cmidrule{2-11} 
& TrafficTL (Ours) & 4.09  & \uline{5.87} & \uline{12.63} & 4.03 & \uline{5.69} & \uline{12.65} & \uline{4.06} & \uline{5.71} & \uline{12.78} \\ 
& TrafficTL (w/o TCB) & 4.49 & 5.97 & 13.49 & 4.55 & 5.88 & 13.86 & 4.59 & 5.97 & 14.18 \\
& TrafficTL (w/o GRB) & 4.45 & 5.85 & 13.42 & 4.43 & 5.73 & 13.48 & 4.49 & 5.80 & 13.82\\
\midrule
& & \multicolumn{9}{c}{Nav-BJ to Nav-HZ}  \\ 
\cmidrule(lr){3-11}  
& & \multicolumn{3}{c}{\SI{15}{\minute}}  & \multicolumn{3}{c}{\SI{30}{\minute}} & \multicolumn{3}{c}{\SI{60}{\minute}}\\
\cmidrule(lr){3-5}\cmidrule(lr){6-8}\cmidrule(lr){9-11}
& &  MAE & RMSE & MAPE  & MAE & RMSE & MAPE & MAE & RMSE & MAPE  \\ \cmidrule(lr){1-11}
\multicolumn{1}{c}{\multirow{6}{*}{Non-transfer}} 
& TGCN & 6.20  & 9.29  & 19.86 & 6.25  & 9.33  & 20.04 & 6.36  & 9.44  & 20.29 \\
& ASTGCN & 6.30  & 9.29  & 20.29 & 6.31  & 9.33  & 20.46 & 6.43  & 9.39  & 20.76 \\
& DCRNN  & 3.74  & 6.49  & 13.90 & 4.03 & 6.72  & 14.69 & 6.15  & 9.10  & 20.77 \\
& TGCN$^\dagger$ & 7.23 & 10.39 & 20.18 & 8.74  & 12.07 & 21.28 & 7.55 & 10.58 & 22.31 \\
& ASTGCN$^\dagger$   & 7.14 & 10.83 & 20.98 & 7.92 & 11.56 & 22.15 & 8.80 & 13.13  & 24.46 \\ 
& DCRNN$^\dagger$    & 4.70  & 6.80  & 14.06 & 5.18 & 7.46 & 15.10 & 7.85 & 11.73 & 22.08 \\
& XGBoost  & 3.95  & 5.90  & 12.27 & 4.08  & 6.05  & 12.65 & 4.23  & 6.21  & 13.07 \\ 
\cmidrule(lr){1-11}  
{\multirow{4}{*}{Transfer}}     & TL\_DCRNN & 6.88 & 9.41  & 18.26 & 8.14   & 10.82 & 24.66 & 11.56 & 16.59  & 30.62 \\ 
& MetaST & 3.42 & 5.01  & 11.83 & 3.52  & 5.13  & 12.17 & 3.65  & 5.28  & 12.60  \\ 
& RegionTrans & 4.20 & 5.55  & 13.87 & 4.45  & 5.89  & 14.23 & 4.69  & 6.26  & 14.94  \\ 
\cmidrule{2-11} 
& TrafficTL (Ours)  & \uline{3.07} & \uline{4.37} & \uline{9.34}  & \uline{3.02}  & \uline{4.25}   & \uline{9.38} & \uline{3.09}  & \uline{4.33} & \uline{9.63}  \\ 
& TrafficTL (w/o TCB) & 4.20 & 5.54 & 11.77 & 4.19 & 5.43 & 11.99 & 4.29 & 5.55 & 12.37 \\
& TrafficTL (w/o GRB) & 4.39 & 5.90 & 12.10 & 3.95 & 5.72 & 12.19 & 4.52 & 5.77 & 12.91\\
\bottomrule
\end{tabular}
    \begin{tablenotes}
        \footnotesize
        \item $^{\dagger}$ The adjacency matrix of these models was acquired by randomly removing edges on the original adjacency matrix following \cite{GCLempirical}.
    \end{tablenotes}

\end{table*}

\begin{table}[t!
]
    \caption{Performance Comparison of Different Areas. Metrics are shown in MAE/RMSE/MAPE. MAPE is Reported in Percentage (\%).}
    \centering
    \begin{tabular}{clcc} 
    \toprule
    \multirow{2}{*}{Approach} & \multicolumn{3}{c}{Nav-BJ to Hongkong} \\ 
    \cmidrule(l){2-4}
    & \SI{15}{min} & \SI{30}{min} & \SI{60}{min}  \\ 
    \cmidrule{1-4}
    \multirow{1}{*}{DCRNN} 
     & $ 2.82 / 4.99 / 6.72 $ & $ 3.18 / 5.47 / 7.80 $ & $ 3.76 / 6.48 / 9.50 $  \\
    \multirow{1}{*}{TrafficTL} 
    & $ 2.27 / 4.57 / 6.54 $ & $ 2.87 / 4.80 / 7.16 $ & $ 3.56 / 5.98 / 9.06 $  \\
    \midrule
     & \multicolumn{3}{c}{Nav-BJ to PEMSD7} \\ 
     \cmidrule{2-4}
     & \SI{15}{min} & \SI{30}{min} & \SI{60}{min}  \\ 
    \cmidrule{1-4}
     \multirow{1}{*}{DCRNN} 
     & $ 2.75 / 4.48 / 6.03 $ & $ 2.97 / 5.15 / 6.78 $ & $ 4.20 / 6.91 / 9.56 $  \\
    \multirow{1}{*}{TrafficTL} 
    & $ 2.05 / 3.90 / 4.17 $ & $ 2.34 / 4.45 / 4.93 $ & $ 2.72 / 5.24 / 5.86 $  \\
    \bottomrule
    \end{tabular}
    \label{tbl:hkpems}
\end{table}
\subsection{Model Performance Comparison}
\label{sec:modelperformcomp}

Table. \ref{tbl:perform} presents the simulation results of the proposed and baseline approaches. Underlines highlight the best-performing results. It is obvious that the proposed TrafficTL outperforms all competing baselines by achieving the lowest MAPE and RMSE. MAE is also the lowest in most cases and is minusculely inferior to the best non-transfer model DCRNN in the two cases. 

Compared with the non-transfer models without $^{\dagger}$, TrafficTL adds additional information from the source city data for the traffic prediction task in the target city.
As a non-transfer approach, T\_GCN, ASTGCN, and DCRNN are trained using one day of the target data training for prediction.
TrafficTL, on the other hand, employs the data from the source city instead of the target city in training, which decreases the inaccuracy caused by defective prior knowledge in the target city's data in comparison.
It effectively transfers potential knowledge from the source city to the target city, reducing negative transfer.
Additionally, TrafficTL is superior to XGBoost in that it can simultaneously learn the data of large cities, allowing it to understand the later state of the target city. XGBoost is capable of learning the temporal features of individual roads from corresponding single nodes, leading to valuable predictions. Specifically, the data used for these predictions is more closely aligned with the distribution of data collected for the target road, allowing for a more accurate representation of traffic conditions on the concrete road. However, the potential for missing data and long intervals between training and test datasets can ultimately hinder the ability of XGBoost to forecast future traffic conditions in a target city accurately.
Compared to the model with $^{\dagger}$ and without, incorrect prior knowledge negatively impacts the whole model. Nevertheless, TrafficTL avoids the risk of negativization initially because it does not rely on information from unreliable data.

In terms of the transfer models, Transfer models are overall better than non-transfer models, except for TL\_DCRNN. The reason for this is that the model’s performance on the target task may be hampered by the generalization gained from the source domain. TL\_DCRNN may result in different types of roads being classified together. Then features from other roads may interfere with learning road features in one class, causing negative effects. Therefore, TrafficTL is taken to learn the respective road section situations, even if the same type of intersection is divided into multiple classes, to minimize the occurrence of misclassification and to try to avoid the interference of negative transfer. 

\begin{figure*}[htbp]
\centering
\begin{minipage}[t]{1.0\textwidth}
\centering
\includegraphics[width=18cm]{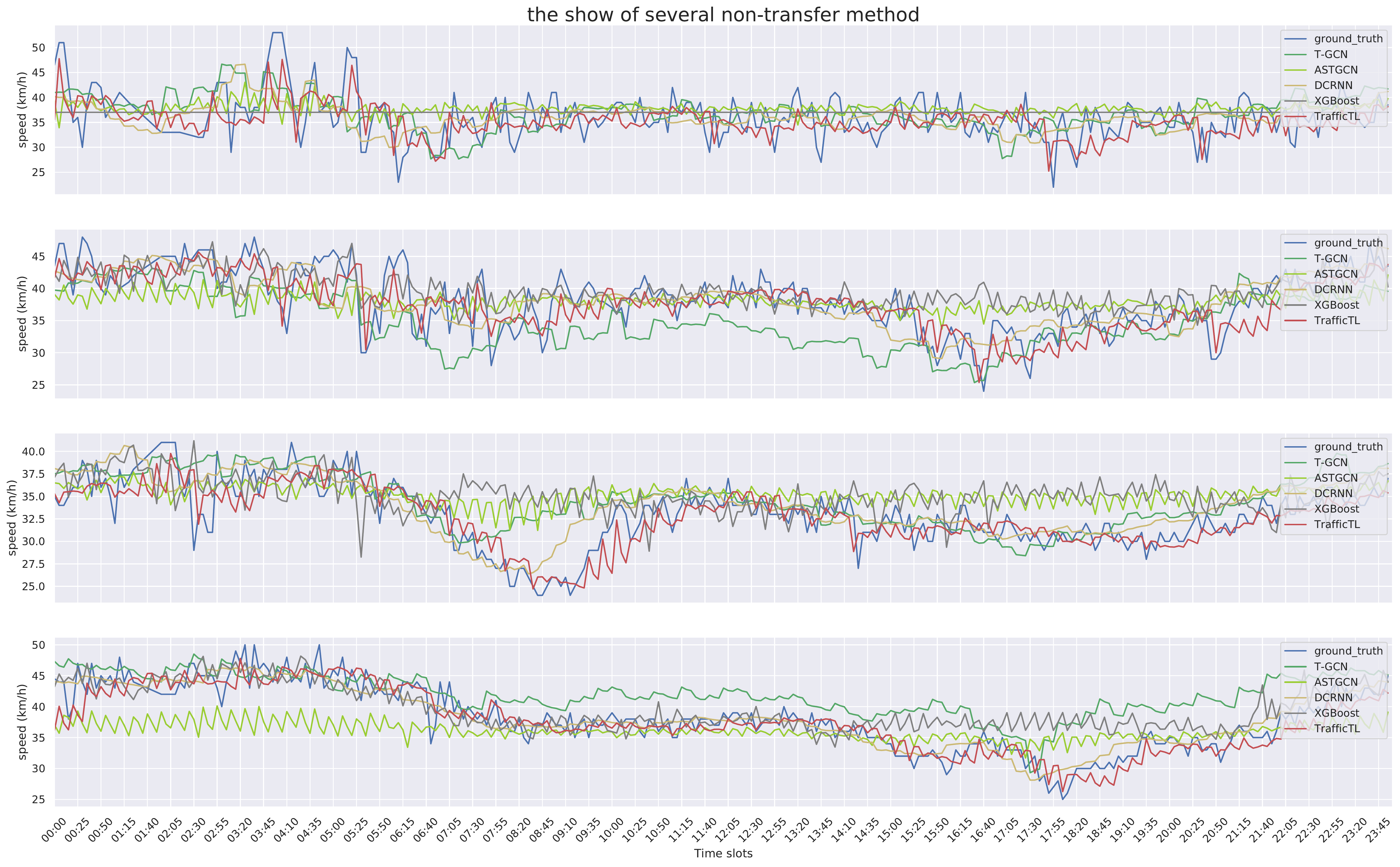}
\caption{The predictions of T-GCN, DCRNN, ASTGCN, and TrafficTL for the selected roads in Nav-HZ.}
\label{fig:traffic_stops}
\end{minipage}
\begin{minipage}[t]{1.0\textwidth}
\centering
\includegraphics[width=18cm]{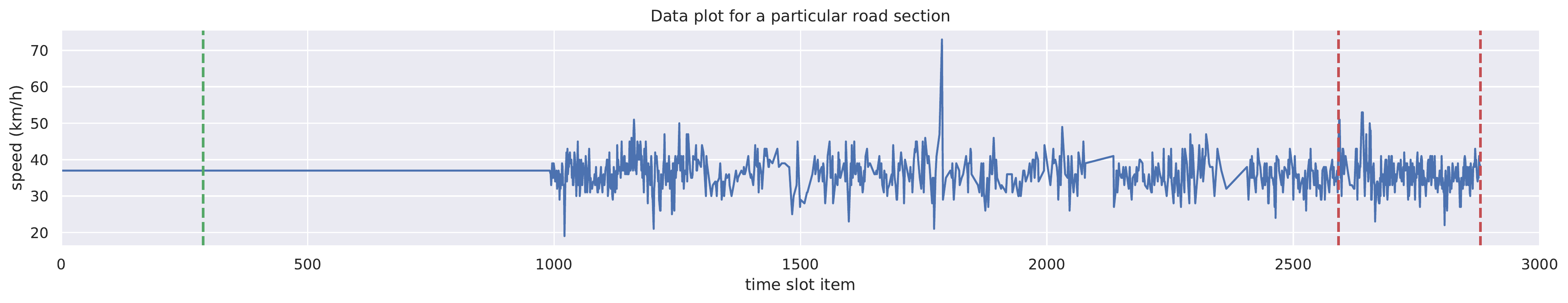}
\caption{An illustration of one road in dataset.}
\label{fig:dataroad}
\end{minipage}
\begin{minipage}[t]{1.0\textwidth}
\centering
\includegraphics[width=18cm]{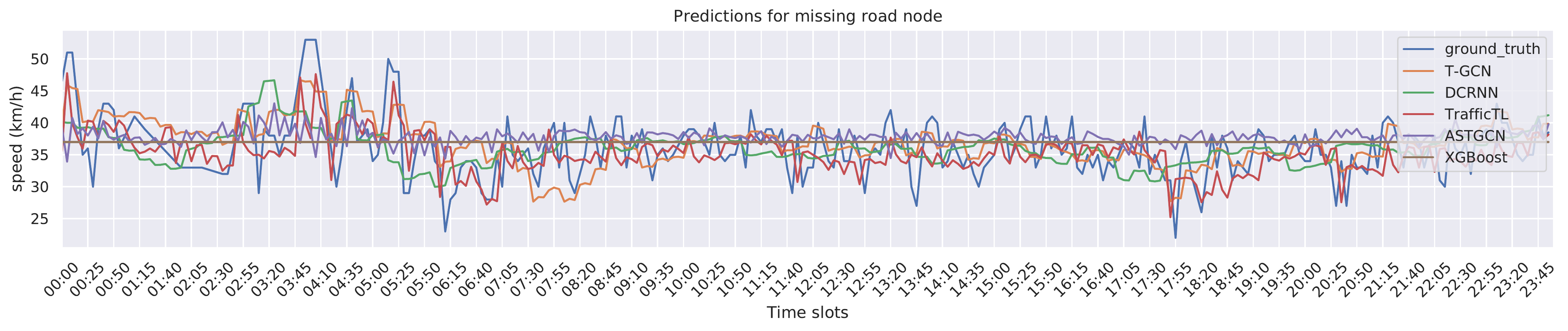}
\caption{Predictions with several methods of the data between red lines.}
\label{fig:missing_pre}
\end{minipage}
\end{figure*}

TrafficTL outperforms all three baselines, namely, TL\_DCRNN, MetaST, and RegionTrans. 
Compared with TL\_DCRNN, TrafficTL uses a more flexible node clustering method  (TCB) and generates a comprehensive dataset to alleviate the source and target dataset gap. 
In addition, an error rectified module (GRB) is applied to eliminate the error between the prior knowledge and the clustered data. 
Compared with MetaST, RegionTrans, we apply ensemble models to distinguish parameters for reducing the conflicts caused by combining coefficients in the direction of gradient descent.
Furthermore, our approach surpasses the non-transfer model in terms of long-term series prediction stability.
When predicting the long-term traffic speed in the target city, a much larger amount of historical data is required to extract the long-term data dependency. 
However, such historical data are scarce in the target city. Incorporating source data in this scenario adds value to long-term prediction, which cannot be utilized in other baselines.
Finally, the dataset with fewer nodes (Nav-HZ) produces superior results when learning from source datasets since the dataset in the source city provides sufficient auxiliary samples for the target with fewer nodes to learn.
Fig. \ref{fig:traffic_stops} visualizes the predicted values of several non-transfer methods for four road sections in Nav-HZ.\footnote{DCRNN results are adopted from DL-Traff-Graph \cite{jiang2021dl}. We employ source codes published by the authors for T-GCN \cite{Zhao2020TGCNAT} and ASTGCN \cite{ASTGCN}.} In order to compare the transfer effects among different countries and datasets, we also used the Hong Kong dataset and the PEMSD7 dataset to compare different types of regions. We use the second-best performing DCRNN in Table \ref{tbl:perform} as the comparison model, and Table \ref{tbl:hkpems} shows the performance of TrafficTL on the two datasets. From Table \ref{tbl:hkpems}, we can see that the transfer effect of TrafficTL is improved on both datasets. The improvements are greater on dataset PEMSD7 than Hongkong, which may be due to the difference in the similarity between the datasets.

\subsection{Ablation Study}
\label{number}
Three components comprise TrafficTL, where TCB enables unsupervised clustering of traffic nodes across multiple cities, and GRB corrects prior and cumulative errors. 
Finally, the ensemble calculates the results across different clusters. 
To evaluate the performance of TCB and GRB on TrafficTL, we remove a block each time while keeping the others.
Note that the ensemble aggregates the results of each cluster, which change according to the clustering module TCB. Hence, it is coupled with TCB in the ablation study.
We define two new variants of TrafficTL as follows:
\begin{itemize}
    \item Traffic(w/o TCB): The traffic nodes of the two cities are not clustered by TCB.
    \item Traffic(w/o GRB): The adjacency matrix is the typical $\operatorname{diag}(\bm{A}^{\mathcal{S}_i}, \bm{A}^{\mathcal{T}_i})\in\mathbb{R}^{N^{\textnormal{C}}_i\times N^{\textnormal{C}}_i}$ instead of the similarity matrix $\bm{W}$ generated by GRB.
\end{itemize}

The results of the ablation study are shown in Table \ref{tbl:perform}. 
Without the TCB or GRB block, the models' performance falls short of TrafficTL.
Nonetheless, TrafficTL(w/o TCB) and TrafficTL(w/o GRB) still perform better in the transfer process from Nav-BJ to Nav-SH and Nav-HZ than the non-transfer models.
This result is caused by the fact that there are still data from the source city that can be used as learning samples for the target city's traffic prediction.
In addition, a comparison of the results obtained from the transfer of traffic data from Nav-BJ to Nav-SH versus the transfer from Nav-BJ to Nav-HZ revealed that the latter process resulted in a more successful transfer when the target city (Nav-HZ) had a smaller number of road nodes (413). Specifically, the transfer from a source city with a larger number of road nodes (e.g., Nav-BJ) to a target city with a smaller number of road nodes (e.g., Nav-HZ) was found to be more successful than the transfer from the same source city to a target city with a larger number of road nodes (e.g., Nav-SH). This suggests that the presence of fewer road nodes in the target city may facilitate the process of finding similar nodes in the source city in terms of traffic states, leading to a more effective transfer. Moreover, it also indicates that better results can be obtained when performing transfer learning in truly small cities. While it is unfortunate that datasets from three large cities were used as ground truth to validate the model's effectiveness, this finding suggests that better results may be obtained when performing transfer learning on truly small cities that have fewer roads.

\subsection{Transfer ability analyze}
In this sub-section, we try to give an example to clarify transferable knowledge. Fig. \ref{fig:dataroad} shows the data collected on a particular road. It can be seen that there are missing values for a significant portion at the beginning of the section. In this experiment, only a tiny portion of the data prior to the green line is utilized as training data, while the portion after the green line is used as test data to evaluate the model's performance.

Fig. \ref{fig:missing_pre} shows the predictions of each model for the ground truth between the two red lines in Fig. \ref{fig:dataroad}.
As demonstrated in Section \ref{sec:modelperformcomp}, XGBoost's performance is largely dependent on the historical data distribution for a specific road. While this focus on individual road data can be beneficial, it also limits XGBoost's ability to make accurate predictions in situations where no available data for learning.
As seen in Fig. \ref{fig:missing_pre}, XGBoost cannot predict the future well from historical values with missing data because it uses historical data for each road segment, i.e., only temporal attributes are involved. Several other models can learn some traffic states of the road node from the normal recorded values in other adjacent road nodes through the adjacency matrix. However, they are limited by the size of the available data, and the performances of these models are not better than the predictions learned by TrafficTL through a large amount of reliable source city data.

\subsection{Parameter Sensitivity}
In this subsection, we investigate the influence of cluster number $c$ and source data quantity $n$ on the transfer prediction performance.
Given the small data size (fewer nodes) of Nav-HZ, it is more representative of the task of providing traffic forecasts for small cities by transferring data from large cities. The results presented in this subsection are based on the transfer from Nav-BJ to Nav-HZ.

\subsubsection{The optimal number of clusters}
We explore the ideal number of clusters in this test. 
Selecting the optimal number of clusters $c$ is a well-known challenging problem in unsupervised clustering \cite{wang2006characteristic}
In TrafficTL, we explore the optimal cluster number according to mutual information and model performance, and the candidate $c$ values are from $2$ to $20$.
Note that mutual information is calculated according to the likelihood that the possibility of the source and target city's nodes (with dataset $X^{\textnormal{C}}$) are classified, which is described in detail in Sec. \ref*{sec:TCB}.

\begin{figure}[!ht]
\centering
\includegraphics[width=\linewidth]{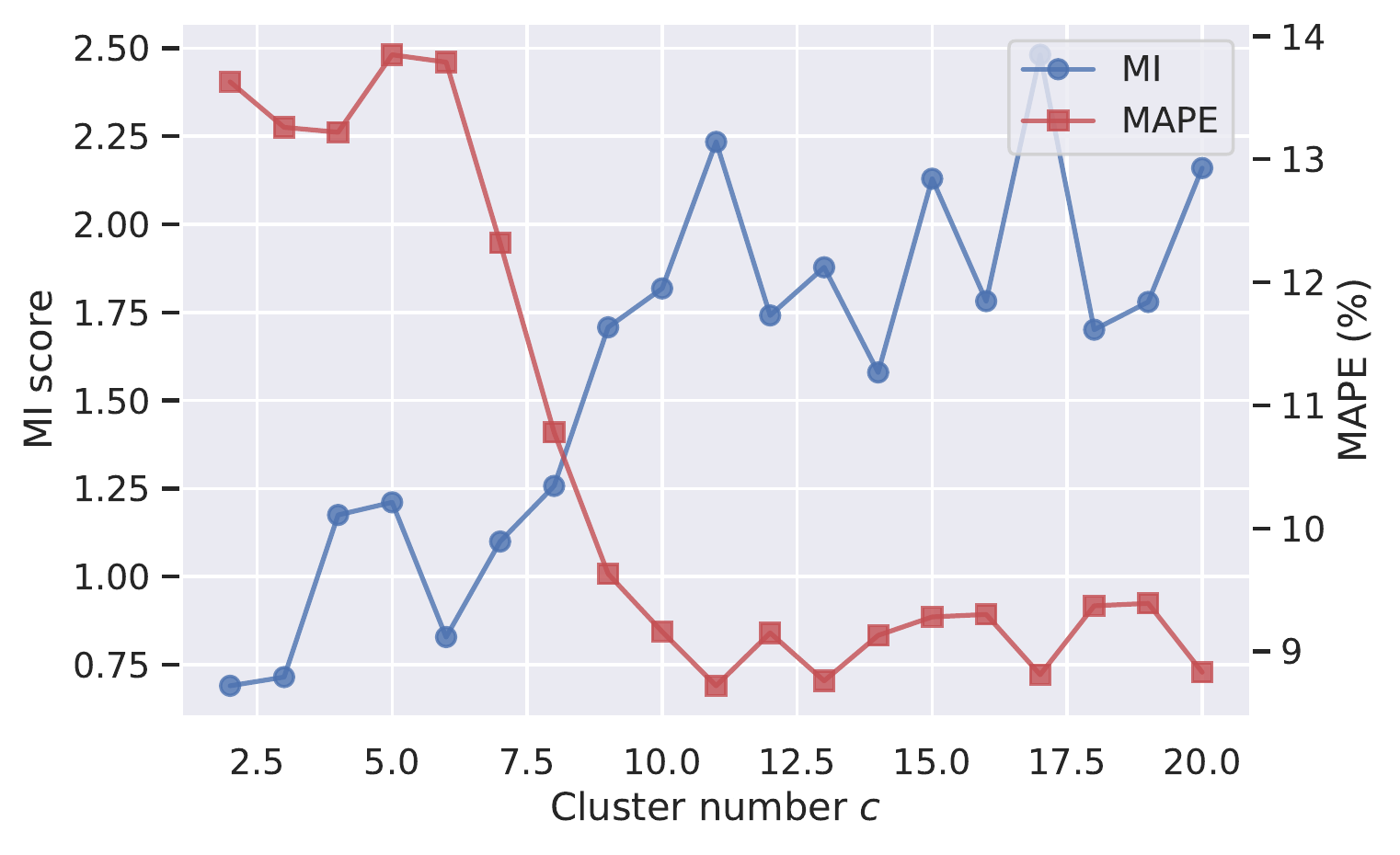}
\caption{The mutual information scores and model prediction performance shown with MAPE under different number of clusters.}
\label{fig:cl_mape_mi}
\end{figure}

The Mutual Information clustering scores (MI scores) of TCB are shown in Fig. \ref{fig:cl_mape_mi}. 
When cluster number $c$ gets large, there is a tendency for the clusters to share more mutual information, which is consistent with the intuition. 
The larger the number of clusters, the more nodes each cluster contains, and the more fine-grained the same features.
However, increasing the number of clusters increases training time and model storage requirements.
From Fig. \ref{fig:cl_mape_mi}, we can observe that when cluster number $c$ reaches $9$, the trend of MI increase is slowed down. 
Besides, the figure also shows the performance (MAPE) of different $c$ in Fig. \ref{fig:cl_mape_mi}.
A negative correlation is clearly reflected between the MI score and MAPE.
Furthermore, the cluster block TCB based on mutual information provides an unsupervised node aggregation method. The correlation between the MI score and MAPE demonstrates its effectiveness.


  
\subsubsection{Transferability of source data quantity}

The concept of transfer learning is to use a large amount of data from other cities to assist cities where data is scarce.
Therefore, a natural question arises: how much source information is required to help the target city learn? Finding an appropriate amount of data minimizes the computational effort of the model and improves its efficiency. We investigate the effect of source-target data quantity size ratio $n$ used on the transfer performance.

\begin{figure}[!t]
\centering
\includegraphics[width=\linewidth]{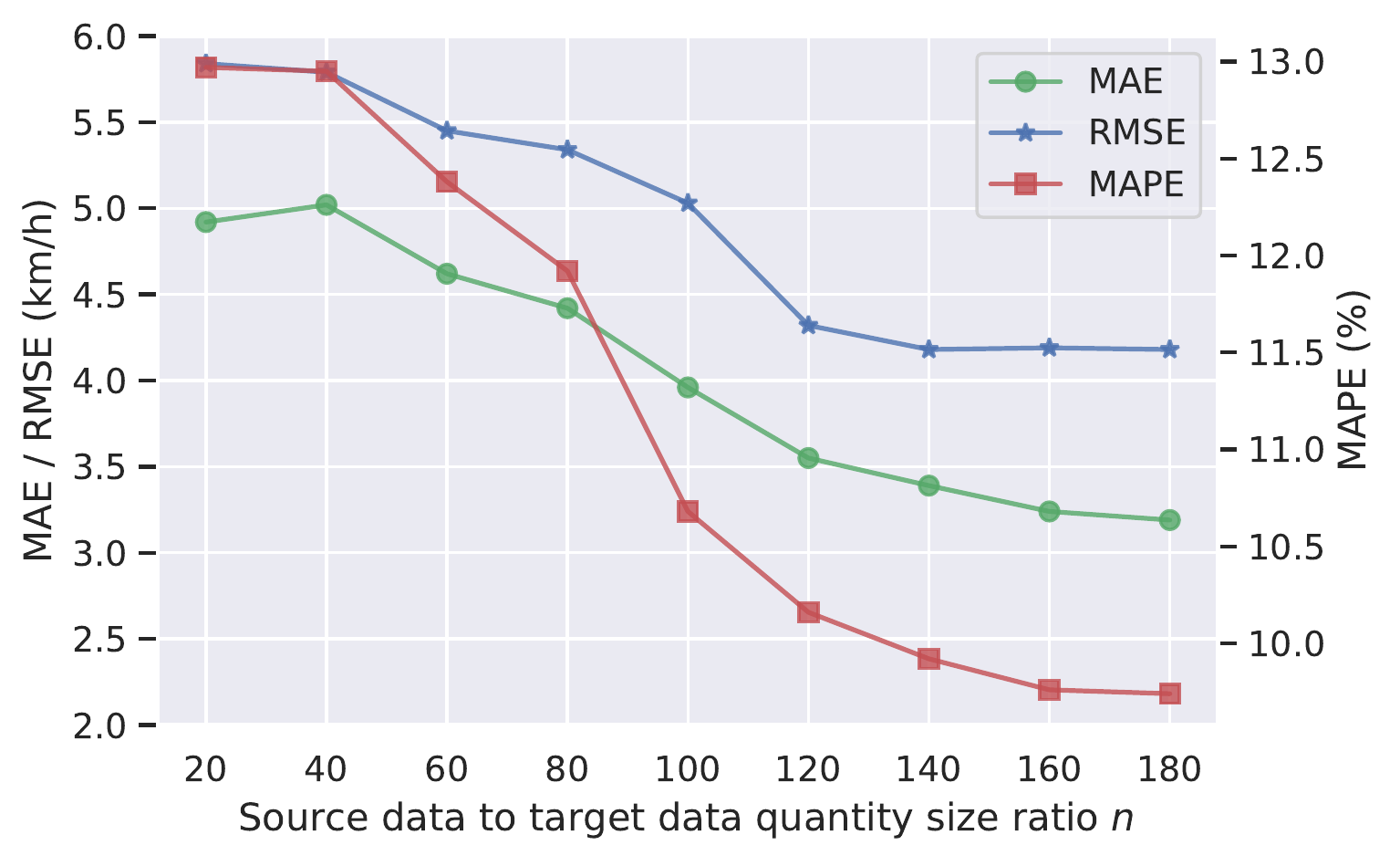}
\caption{The performance metrics under various ratio of source-target data quantity size.}
\label{fig:ratio_mape}
\end{figure}

As illustrated in Fig. \ref{fig:ratio_mape}, MAE and RMSE decrease before $n = 120$, where the degradation is slowed down thereafter. 
When the source data ``richness'' is saturated, more source data do not further improve the transferability, thereby, the performance metrics.
In addition, increasing $n$ involves more training data. However, the figure illustrates that increasing $n$ does not lead to a consistent linear improvement in performance, resulting in unnecessary computation.

\section{Discussion} \label{sec:discussion}
\textcolor{black}{In this section, we aim to explore potential concerns related to the applications of our work. As mentioned earlier, during the initial stages of urban development, the deployment of equipment can be quite expensive. Our previous sections have highlighted that transfer learning can potentially reduce these costs by offering cost-effective city-wide traffic forecasts. However, two questions naturally arise with this model: ``Is it necessary to build a data collection network of fixed-point devices for cities?" and ``Why do we need transfer learning?"}
 
\textcolor{black}{
Regarding the first problem, Floating Car Data (FCD) has been widely utilized in transportation research for some time. According to a 2008 report \cite{Rev1}, FCD offers low cost and ease of access, and with improvements in GPS accuracy, it has become a potentially viable data source for traffic analysis. However, several academic papers have noted challenges in utilizing FCD as a source of information for traffic analysis. These challenges include privacy concerns when working with companies that collect FCD, and difficulties in effectively integrating and processing FCD data collected by the government.}
\textcolor{black}{
\begin{itemize}
    \item One such challenge is privacy concerns when working with companies that collect FCD. As FCD collection becomes more feasible with advancements in GPS technology \cite{Rev5}, there is growing concern about protecting personal privacy \cite{Rev8, Rev11}. When working with major providers of FCD collection sources such as Didi and Uber, governments may become embroiled in commercial transactions and privacy-related issues \cite{Rev4}.
    \item Another challenge is related to data collection by the city itself. There may be difficulties in effectively integrating and processing data when the government collects FCD. Several challenges arise when collecting FCD data, such as inaccurate data resulting from drivers forgetting to end their trips, the complexity of integrating multiple tracks of varying lengths for a particular road \cite{Rev4, Rev5, rev6, Rev9}, and the limitation of data coverage, as the government can only mandate the installation of FCD collection devices for its controlled departments such as taxi groups, rather than requiring all residents to upload their data \cite{Rev8}.
\end{itemize}}
\textcolor{black}{Further, we figured that the form of data used has a high correlation with the task and the proposed method according to \cite{jiang2021dl}. Methods based on graph structures have become more prevalent since the publication of STGCN in 2018, as fixed-point data in the Roadside Collected Data (RCD) style makes it easier to turn traffic networks into graphs that capture spatial information. While 58\% of the works still relied on RCD as their data source, 42\% utilized FCD as their data source \cite{Rev2}. City managers aiming for growth should consider utilizing both types of data \cite{Rev3} and give greater attention to establishing their own intelligent transportation data collection networks.
}
\begin{figure}
    \centering
    \includegraphics[scale=0.2]{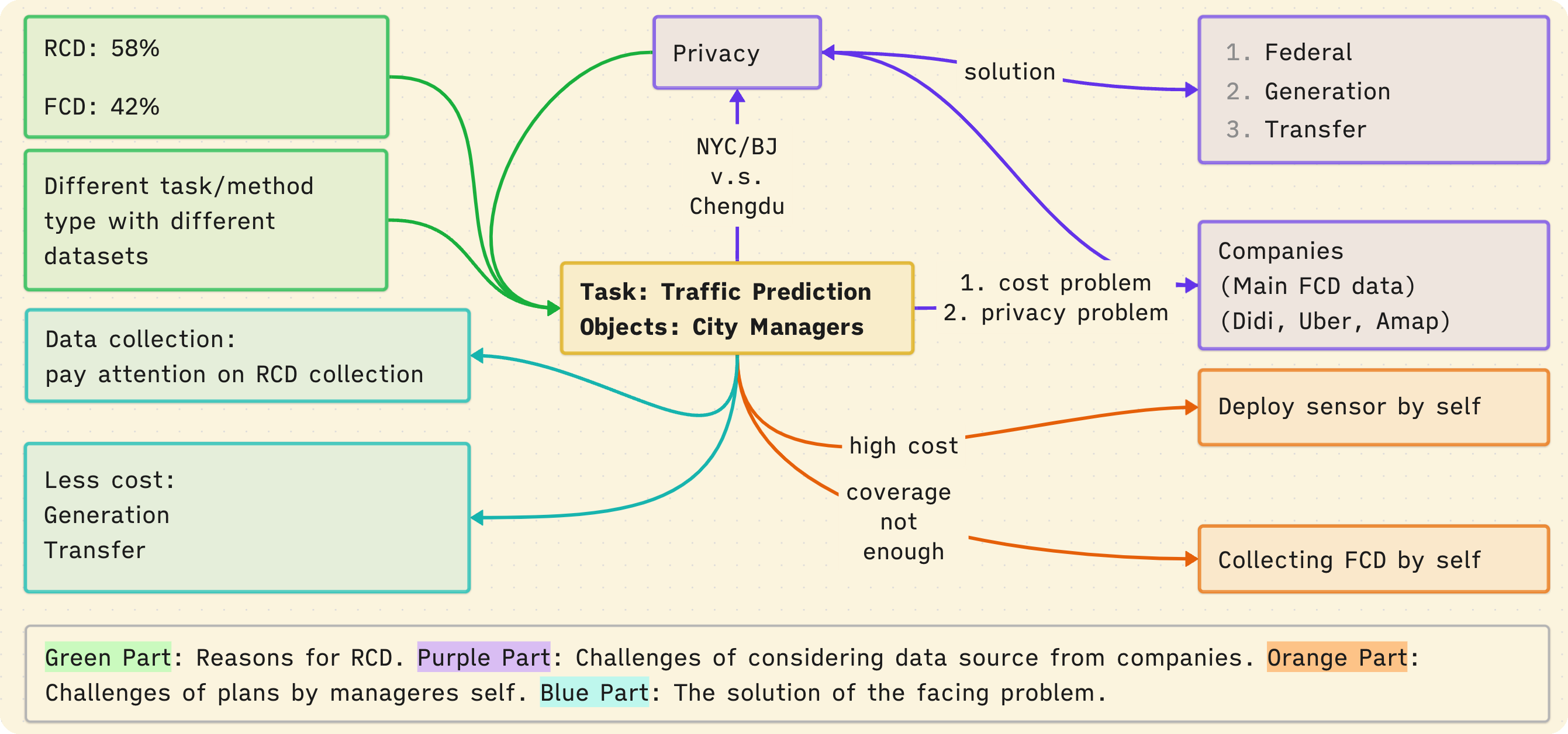}
    \caption{(best viewed
in color) Green Part: Reasons for RCD. Purple Part: Challenges of considering data source from companies. Orange Part: Challenges of plans by managers self. Blue Part: The solution of the facing problem.}
    \label{fig:discuss}
\end{figure}

\textcolor{black}{Regarding the second problem, data collection can often take several months \cite{Rev5}. Utilizing transfer learning as a means to expedite the implementation of urban prediction models is a viable approach that only entails a modicum of additional expenses \cite{Rev13}. The deployment of detectors by the government is not a one-time affair, and implementing it in phases would prolong the data collection period and extend the time required to build and deploy models. Privacy concerns restrict the public release of collected data, making transfer learning a promising approach to overcome these limitations. Models trained on large datasets from cities may perform poorly on smaller datasets \cite{Rev13}, leading to the exploration of alternative methods such as federal learning and data generation techniques. Transfer learning, which utilizes large amounts of publicly available data to optimize models, is a promising approach that reduces the time and financial burden associated with constructing models during the early stages of urban development.}

\textcolor{black}{In conclusion, transfer learning is a cost-effective and beneficial approach for city managers, particularly for those cities in their early stages of development. We have distilled the potential risks, challenges, and questions into Figure \ref{fig:discuss}, offering a visual illustration of the concerns.}

\section{Conclusion}\label{section:VI}
In this paper, we propose a cross-city traffic prediction method named TrafficTL. 
This approach can extract common knowledge in datasets from data-rich cities to provide traffic forecasts for data-scarce cities and help traffic management agencies make decisions accordingly.
TrafficTL devises a node aggregation scheme (TCB) to reduce the negative impact of the data distribution difference between source and target cities. 
A graph reconstruction method (GRB) is proposed to rectify the negative impact of prior knowledge inaccuracies from data-scarce cities and the accumulated errors that come from unsupervised clustering in TCB. 
Additionally, it integrates ensemble learning to distinguish model parameters to lessen the adverse effects of parameter sharing across node clusters. 

To evaluate the performance of TrafficTL, we conduct comprehensive experiments based on three real-world datasets. Comparisons with the state-of-the-art baselines show the superiority of TrafficTL over existing solutions.
In addition, we test the model performance over a variety of cluster numbers to reveal the influence of clustering on the transferability of TrafficTL. Furthermore, we investigate the ratio of the source data provision in transfer to reduce the computational burden and trial-and-error cost.

In the future, we plan to explore other transfer learning mechanisms from perspectives such as traffic graph construction and graph comparison. We look forward to follow-up research on developing general schemes for traffic transfer prediction based on graph learning. In addition, the findings of our analysis in Section \ref{number} indicate that transfer learning may be effective for small cities with fewer urban roads. However, the data used in our study were drawn from three large cities in order to provide ground-truth for our experiments, which may not be representative of the transfer effects between large and small cities. We look forward to obtaining more consistent data in order to validate our results in the future.


\bibliographystyle{IEEEtran}
\bibliography{IEEEabrv, ref}

\ifCLASSOPTIONcaptionsoff
  \newpage
\fi

\end{document}